\documentclass[10pt,journal,compsoc]{IEEEtran}
\usepackage{amsmath,amssymb,amsfonts}
\usepackage{algorithm}
\usepackage{algpseudocode}
\usepackage{multirow}
\usepackage{latexsym}
\DeclareMathOperator*{\argmax}{argmax}
\usepackage{graphicx}
\usepackage{subfigure}
\usepackage{textcomp}
\usepackage{bbding}

\newcommand{\tabincell}[2]{\begin{tabular}{@{}#1@{}}#2\end{tabular}}

\renewcommand\tabcolsep{3.0pt}

\ifCLASSOPTIONcompsoc
  \usepackage[nocompress]{cite}
\else
  \usepackage{cite}
\fi

\ifCLASSINFOpdf

\else

\fi

\begin{document}
%
\title{Adversarial Examples Versus Cloud-based Detectors:
A Black-box Empirical Study}
%
%
%
%

\author{
	Xurong Li,
        Shouling Ji,
         Meng Han,
        Juntao Ji,
        Zhenyu Ren,
        Yushan Liu,
        and Chunming Wu
\IEEEcompsocitemizethanks{\IEEEcompsocthanksitem X. Li, S. Ji, J. Ji, Z. Ren and C. Wu are with the Institute of Cyberspace Research and the College of Computer Science and Technology at Zhejiang University, Hangzhou, Zhejiang, 310027, China; S. Ji is also with the School of Computer Science and Technology at Georgia Institute of Technology, Atlanta, Georgia 30302, USA.
\protect\\
E-mail: \{lixurong, sji, 3160102420, rzheny, wuchunming\}@zju.edu.cn
\IEEEcompsocthanksitem M. Han is with the College of Computing and Software Engineering, Kennesaw State University, Marietta, GA, 30060.
\protect\\E-mail: mhan9@kennesaw.edu.
\IEEEcompsocthanksitem Y. Liu is with the Department of Electrical Engineering, Princeton University, Princeton, NJ, 08540.
E-mail: yushan@princeton.edu.

\IEEEcompsocthanksitem This paper was accepted by IEEE Transactions on Dependable and Secure Computing (TDSC).

}
}

\IEEEtitleabstractindextext{%
\begin{abstract}
Deep learning has been broadly leveraged by major cloud providers, such as Google, AWS and Baidu, to offer various computer vision related services including image classification, object identification, illegal image detection, etc.
While recent works extensively demonstrated that deep learning classification models are vulnerable to adversarial examples, cloud-based image detection models, which are more complicated than classifiers, may also have similar security concern but not get enough attention yet.
In this paper, we mainly focus on the security issues of real-world cloud-based image detectors. Specifically, (1) based on effective semantic segmentation, we propose four attacks to generate semantics-aware
adversarial examples via only interacting with black-box APIs; and (2) we make the first attempt to conduct an extensive empirical study of black-box attacks against real-world cloud-based image detectors.
Through the comprehensive evaluations on five major cloud platforms: AWS, Azure, Google Cloud, Baidu Cloud, and Alibaba Cloud,
we demonstrate that our image processing based attacks can reach a success rate of approximately 100\%, and the semantic segmentation based attacks have a success rate over 90\% among
different detection services, such as violence, politician, and pornography detection.
We also proposed several possible defense strategies for these security challenges in the real-life situation.

\end{abstract}

}

\maketitle

\IEEEdisplaynontitleabstractindextext

%
\IEEEpeerreviewmaketitle

\IEEEraisesectionheading{\section{Introduction}\label{sec:introduction}}

%
%
%
%
\IEEEPARstart{T}{aking} advantage of the availability of big data and the strong learning ability of neural networks, deep learning outperforms many other traditional approaches in various computer vision tasks such as image classification, object detection, and image segmentation.
Since deep learning often requires massive training data and lengthy training time, many cloud service providers (such as Google, AWS, Baidu, Alibaba, Azure) offer deep learning Applicant Program Interfaces (APIs) for their clients to accomplish computer vision tasks without the need to train models and own a big amount of data.
These APIs can help cloud service users check images for both commercial and non-commercial purposes. For example, the search engine giant Google\footnote{https://cloud.google.com} and Baidu\footnote{https://ai.baidu.com} allow their APIs to identify the category of pictures (e.g., dog, cat); Alibaba Cloud\footnote{https://www.alibabacloud.com} and Azure\footnote{https://azure.microsoft.com} provide APIs to check whether the images are illegal (e.g., pornographic, violent).

However, deep learning has been recently found extremely vulnerable to adversarial examples, which are carefully-constructed input samples that can trick the learning model into producing incorrect results.
Hence, the study of adversarial examples in deep learning has drawn increasing attention among the security community \cite{akhtar2018threat}.
In general, in terms of applications, the research of adversarial example attacks against cloud vision services can be grouped into three main categories: \emph{self-trained classifier} attacks, \emph{cloud-based classifier} attacks, and \emph{cloud-based detector} attacks, as shown in Table \ref{serviceattack} (in this paper, classifiers and detectors refer to the deep learning models in the image field).
An image classifier utilizes training images to understand how a given input variable relates to a class. By contrast, an image detector identifies the bounding areas that are ``worth labelling" in an input image, and then generates a label for each area.
For \emph{self-trained classifiers}, clients upload the training data themselves and attackers know the distribution of the training data in advance \cite{papernot2017practical,hayes2017machine,chen2017zoo}.
For \emph{cloud-based classifiers}, the cloud providers train the classifiers themselves
(e.g., image classifiers on AWS). Attackers do not have the prior knowledge and state-of-art attacks are achieved by making hundreds of thousands of queries to successfully generate an adversarial example \cite{liu2016delving,brendel2017decision}.
For \emph{cloud-based detectors}, the cloud providers train the detectors themselves and integrate the detectors into their detection services, e.g., the detection services for violent and pornographic images provided by Google.
Despite a ``classifier'' is incorporated in the last step, \emph{cloud-based detectors} usually need to contain other modules, such as object detection, image segmentation, and even human judgment in some complicated situations. 
Attacking cloud-based image detectors is a challenging task since it is hard to bypass these complicated techniques simultaneously to launch a successful attack with limited queries.

\begin{table}[!tp]

\centering
\caption{Cloud-based vision services and attacks.}\label{serviceattack}
\begin{tabular}{cc}
\hline

\hline

\hline
Vision Model Service & Attack Method\\
\hline

\hline

\hline
self-trained classifier & Substitution model\cite{papernot2017practical}, MLaaS\cite{hayes2017machine}, ZOO\cite{chen2017zoo}\\
cloud-based classifier & Boundary attack \cite{brendel2017decision}, Ensemble models \cite{liu2016delving}\\
cloud-based detector & Work in this paper\\
\hline

\hline

\hline
\end{tabular}
\end{table}

Although \emph{cloud-based detectors} are playing an increasingly important role, there is still limited work exploring the possibility of adversarial example attacks against the detection of cloud vision services.
A few of the most recent works that paid attention to fool image detectors focused on the standard object detection algorithm \cite{lu2017adversarial}. 
However, such attacks cannot be readily applied to the cloud environment. This is because (i) cloud-based detectors contain several complicated
 modules (object detection, image segmentation, human judgment, etc.), and (ii) cloud-based detectors are presented as black-box to their adversaries.
In \cite{tramer2016stealing}, Florian et al. succeed to steal a machine learning model via public APIs by an equation-solving method. However,
it is impractical to attack a commercial model with hundreds of millions of parameters by a simple equation-solving method. Furthermore, \cite{tramer2016stealing} succeeds to steal a simple model trained by themselves and it belongs to \emph{self-trained classifiers} attack.

To fill this emerging gap, in this work, we take the first step to present attacks on \emph{cloud-based detectors}.
In order to conduct a comprehensive study, we consider the image detector services on five major cloud platforms worldwide including Baidu Cloud, Google Cloud, Alibaba Cloud, AWS and Azure.
In the rest of this paper, we use ``Google'', ``Alibaba'', and ``Baidu'' for short, to present the corresponding cloud services provided by these companies.

In this study, 
by incorporating the image semantics segmentation, we propose
four black-box attack methods on cloud-based detectors, which do not need prior knowledge of
detectors and more importantly, can be achieved within very limited queries.
Specifically, we present the Image Processing (IP) attack, Single-Pixel (SP) attack, Subject-based Local-Search (SBLS) attack and Subject-based Boundary (SBB) attack.
Our empirical study demonstrates that the proposed attacks can successfully fool the cloud-based detectors deployed on the major cloud platforms with a remarkable bypass rate even approaching 100\% as shown in Table \ref{results}.
We summarize our main contributions as follows:\\

\begin{itemize}

\item To the best of our knowledge, this is the first work to study black-box attacks on cloud-based detectors without any access to the training data, model, or any other prior knowledge.
We investigate the components of detectors and develop the black-box attacks on the detectors beyond the previous classifier fooling.

\item We propose four attack methods by incorporating semantic segmentation to achieve a high bypass rate with a very limited number of queries. Instead of millions of queries in previous studies, our methods find the adversarial examples using only a few thousands of queries.

\item We conduct extensive evaluations on the major cloud platforms worldwide. The experimental results demonstrate that all major cloud-based detectors can be bypassed successfully by our attacks. All the tests only rely on the APIs of these cloud service providers, which verify the feasibility of our proposal in practice.

\item We discuss the potential defense solutions and the security issues. By revealing these vulnerabilities, we provide a valuable reference for academia and industry for developing an effective defense against these attacks.
We reported the vulnerabilities to the involved cloud platforms and received very active and positive acknowledgements from them.

\end{itemize}

\begin{table}[!tp]
\centering
\caption{Success rates of cloud-based detectors attack.}\label{results}
\resizebox{3.5in}{!}{
\begin{tabular}{cccccc}
\hline

\hline

\hline
Platforms&\tabincell{c}{Detection Services}&\tabincell{c}{ IP }&SP &SBLS & SBB \\
\hline

\hline

\hline
\multirow{4}*{Baidu}
&violence  &100\%&88\%&100\%& 91\%\\
&politician &100\%&96\%&60\%&82\%\\
&pornography  &100\%&13\%&34\%&46\%\\
\hline

\hline
\multirow{2}*{Google}
&violence &100\%&8\%&--&75\%\\
&pornography &100\%&59\%&--&78\%\\
 \hline

 \hline
\multirow{3}* {Alibaba}
&violence  &100\%&49\%&72\%&67\% \\
&politician  &100\%&86\%&46\%&67\%\\
&pornography &100\%&12\%&19\%&36\% \\
 \hline

 \hline
 \multirow{2}*{AWS}
&politician &100\%&84\%&--&--\\
&pornography &100\%&50\%&--&--\\
 \hline

 \hline
Azure
&pornography &100\%&91\%&54\%&80\% \\
\hline

\hline

\hline

\end{tabular}
}
\end{table}

\textbf{Roadmap} In the rest of the paper, we begin with the preliminary in Section 2, followed by the threat model and criterion in Section 3.
Section 4 describes the details of our attack algorithms. Section 5 shows experimental results on the cloud-based detectors.
The effects of these attacks and potential defense methods are discussed in Section 6.
Section 7 summaries the related work. Finally, Section 8 concludes this paper and discusses further work.

\section{Preliminary}
\subsection{Neural Networks and Adversarial Examples}
\begin{figure}[htbp]
\centering

\includegraphics[width=1\linewidth]{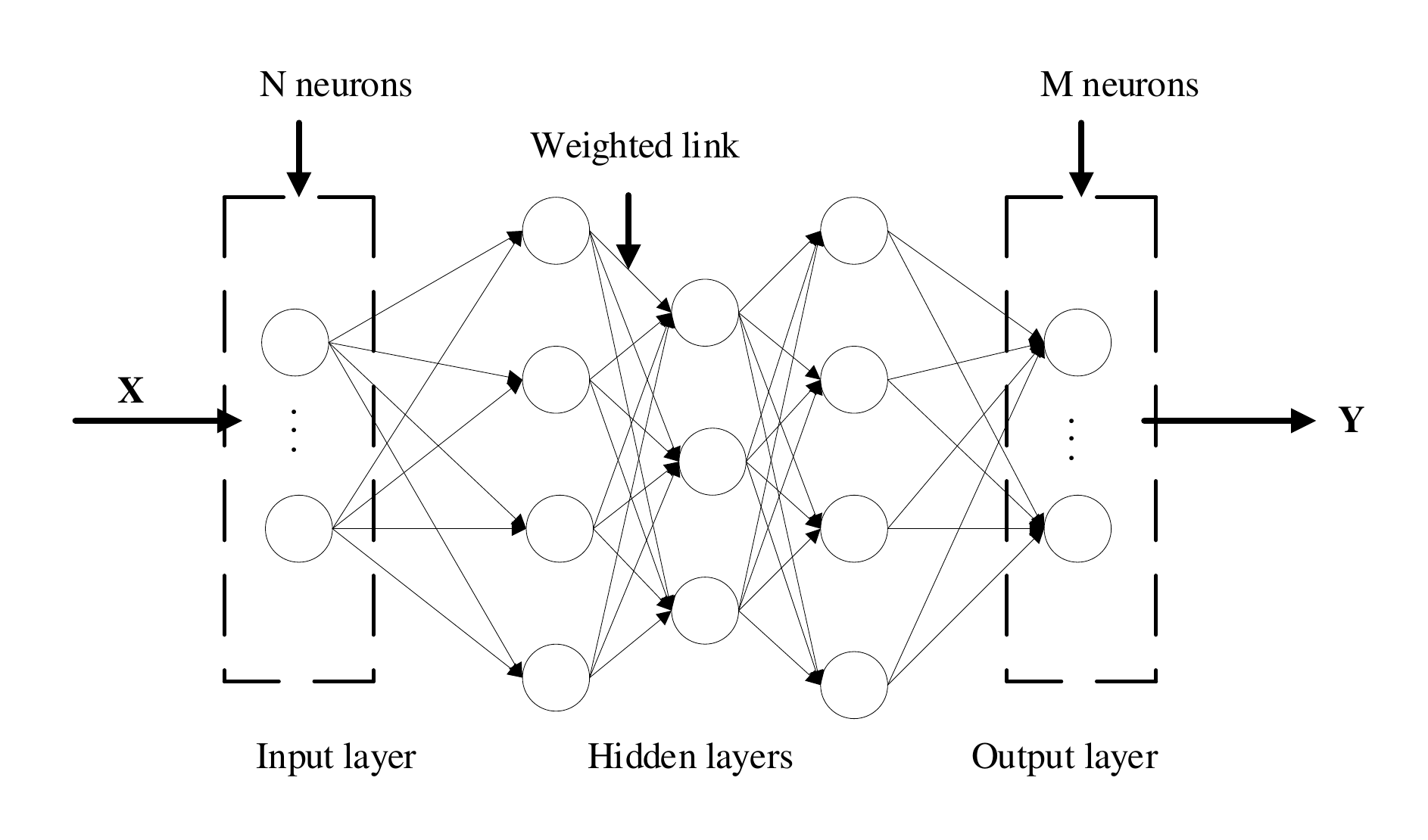}

\caption{A neural network example.}\label{NN}
\end{figure}

A neural network is a function $F(X) = Y$ that accepts $X\in\mathbb{R}^n$ and outputs $Y\in\mathbb{R}^m$, where the model is an $m$-class classifier,
$\mathbb{R}$ is the set of real numbers,
$n$ is the dimension of $X$, and $F$ is the combination of model parameters $\theta$. In this paper, the parameters of the target model are unknown.
The output $Y$ is an $m$-dimensional vector $<y_1, y_2, \cdots, y_m>$, where $y_i$ is the probability of each class, $0 \leq y_i \leq 1$, for $1\leq i \leq m$ and
$\sum_{i=1}^{m}{y_i}=1$. We show the architecture of a Deep Neural Networks (DNN) model in Fig. \ref{NN}.
The final label $ L(X) =  \mathop{\argmax}\limits_{i}{(y_i)}$.
Sometimes in response to a query, cloud models only return a confidence score instead of a probability distribution. Note that there is no correlation between scores in different classes.
For instance, Alibaba Cloud only returns a confidence score from 0 to 100 when being queried.
Probability can leak more information than scores due to the strong correlation between the probabilities and the classes.
Our algorithms could be adapted well in either of the cases (probabilities or scores).

Adversarial example attacks on neural networks were first proposed by Szegedy \cite{szegedy2013intriguing},
wherein well-designed input samples called \emph{adversarial examples} are constructed to fool the learning model. Specifically, adversarial examples are generated from benign samples by adding small perturbation that is imperceptible to human eyes, i.e., $X^* = X+\delta_X$ and $F(X^*) = Y^*$, where $X^*$ is an adversarial example, $\delta_X$ is the perturbation, and $Y^*$ is the adversarial label.
Prior works \cite{stallkamp2012man,rosenberg2017generic} have shown that adversarial examples are detrimental to many systems in the real world.
For instance, under adversarial example attacks, the automatic driving system may take a stop sign as an acceleration sign \cite{stallkamp2012man} and
malware can evade the detection systems \cite{rosenberg2017generic}.
Depending on whether there is a specified target for misclassification, adversarial example attacks can be categorized into two types, i.e., targeted and untargeted attacks.
Since we only intend to make the API service generate an incorrect label, in this paper, we focus on the untargeted attacks.

\subsection{Cloud Vision APIs Based on Neural Networks}
Due to the high cost of storing massive data and intensive computational resource usage of training neural networks for computer vision, it is common in recent years for individuals and small businesses to use cloud platforms to train and perform deep learning tasks.
The cloud service providers normally own plenty of data and computing power, and they actively provide their users with multiple APIs of pre-trained neural networks as their services.
By leveraging these cloud-based classification and detection services, a small fee could allow an application to complete a relatively complicated computer vision task.
We list the computer vision services provided by several major cloud service providers in the market along with
the fees they charge for 1000 queries in Table \ref{fee}.

\begin{table}[!htbp]
\centering
\caption{Services and fees of different cloud vision APIs.}\label{fee}
\begin{tabular}{cccc}
\hline

\hline

\hline
 Cloud Platform&Classifier & Detector &Fee/Query \\
 \hline

 \hline

 \hline
 Baidu & Y & Y&0.17\$/1000 \\

 Google & Y  & Y&1.5\$/1000 \\

 Alibaba & Y &Y&0.27\$/1000\\

 Azure &Y &Y &1\$/1000 \\

AWS &Y &Y &1\$/1000 \\

\hline

\hline

\hline
\end{tabular}
\end{table}

Both classification and detection modules are provided by these computer vision APIs.
The classification module has many applications such as logo recognition, celebrity recognition, animal recognition, etc.
The Detection module aims to find illegal images which violate the content security policy. A detector is more complex than a classifier, as it involves more components such as object detection and image segmentation. Warnings are generated by cloud-based detectors when the outputs of models exceed a threshold.
In this paper, we select the most representative image detection topics, including violence, politics, and pornography.
All experiments are completed only using the free quota provided by these cloud service providers.
This implies that anyone can launch a successful attack by our models in the real world with a very low cost.

\subsection{White-box and Black-box Attacks}
Recent research has proposed several attacks on deep learning models.
Based on the prior knowledge possessed by attackers, adversarial example attacks can be classified to white-box and black-box attacks, as shown in Table \ref{attacktype}, where
\emph{Architecture} means the parameters of a model, \emph{training tools} mean the training methods used when training the model, and
\emph{Oracle} means whether the model gives an output when queried with an input.
\begin{table}[!htbp]
\centering
\caption{Comparison of the prior knowledge of white-box attacks and black-box attacks.}\label{attacktype}
\begin{tabular}{ccccc}
\hline

\hline

\hline
Attack types & Architecture & Training tools & Train data & Oracle\\
\hline

\hline

\hline
white-box &\Checkmark &\Checkmark & \Checkmark& \Checkmark\\
black-box &\XSolid &\XSolid &\XSolid & \Checkmark \\

\hline

\hline

\hline
\end{tabular}
\end{table}

In this paper, we only focus on the black-box attacks against deep learning models, which is even more challenging due to the limited access to the model.
In fact, based on the block-box attacks, it is straightforward to design the white-box attacks.

\subsection{Image Semantic Segmentation}
Semantic segmentation of images includes dividing and recognizing the contents in the image automatically.
Semantic segmentation processes an image at the pixel level, thus we can assign each pixel in the image to an object class.
With the proposal of a full convolutional neural network \cite{long2015fully},
deep learning has been widely adopted in the field of semantic segmentation \cite{badrinarayanan2015segnet}\cite{chen2014semantic}\cite{lin2017refinenet}\cite{peng2017large}.

\begin{figure}[htbp]
\centering
\begin{tabular}{cc}
\includegraphics[width=0.4\linewidth]{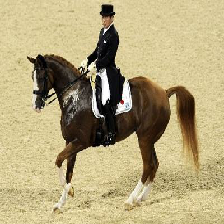}&
\includegraphics[width=0.4\linewidth]{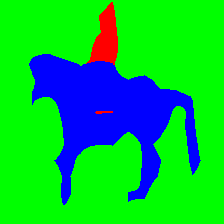}\\
(a) & (b)
\end{tabular}
\caption{Image semantic segmentation example. (a) is the original image and (b) is the semantic segmentation of (a).
There are two classes in the original picture: \emph{person} and \emph{horse}.}\label{segment}
\end{figure}

Through semantic segmentation techniques, we can focus on the \emph{key pixels} of an image, where \emph{key pixels} mean the important pixels contributed to the classification result. If we perturb the \emph{key pixels}, the attacks tend to be easy to implement.
Therefore, the general idea of our attack is using the semantic segmentation model to identify the pixels of an image, and then perturb the pixels of a particular class.
In this paper, we choose Fully Convolutional Networks (FCN)\cite{long2015fully} as the semantic segmentation model, since FCN is one of the most
classical models in the semantic segmentation filed and is sufficiently good for us to conduct our attacks.
Different from the classic CNN, which uses a fully connected layer in the final to obtain a fixed-length feature vector for classification, FCN can take input an image with arbitrary size and produce the corresponding sized output with efficient inference and learning. FCN uses the deconvolution layer to upsample the feature map of the last convolutional layer, restoring it to the same size of the input image.
Then, pixel-by-pixel classification is performed on the upsampled feature map.
For an input $X$, support that $P(i,j)$ is the pixel of $X$ at the location $(i,j)$, $C(P(i,j))$ is the class $P(i,j)$ belongs to, and
$S$ is the subject pixel set of the input. For instance, in Fig. \ref{segment}, we take a \emph{person} as the \emph{subject class}. Thus, we have
$S =subject(X)=\{P(i,j):C(P(i,j))=person\}$.
Similarly, we can get the set of animal images.
The details of the image semantic segmentation are shown in Table \ref{subjectclass}.

\begin{table}[!htbp]
\centering
\caption{Subject class of different images.}\label{subjectclass}
\begin{tabular}{ccccc}
\hline

\hline

\hline
\tabincell{c}{Images \\Type} &Violence&Politician&Pornography  \\
\hline

\hline

\hline
\tabincell{c}{Subject\\ Class}&person&\tabincell{c}{personal \\ face}&person\\

\hline

\hline

\hline
\end{tabular}
\end{table}

In our experiment, the results show that
the perturbation based on semantic segmentation could speed
up the generation of adversarial examples in the cases of violence, politician and pornography.

\section{Threat Model and Criterion }

\subsection{Threat Model}
In this paper, we assume that the attacker is a client and s/he can only access the cloud APIs as a black box.
The only data the attacker can collect is the feedback from the cloud API by the query. Moreover, the attacker can only access the API with a limited number of queries
since it is inefficient and impractical to conduct a large number of queries on cloud platforms.

\subsection{Criterion and Evaluation}\label{metrics}
The goal of adversarial example attacks against detection is to mislead the detector into misclassification.
Nina et al. \cite{shiva2017simple} proposed the concept of top-$k$ misclassification, which means the network ranks the true label below at least $k$ other labels.
In practice, the cloud-based detectors usually produce a label (e.g., violent, pornographic, etc.) after processing an image, which can be used by websites to judge the legitimacy of the image.
Consequently, we choose the top-1 misclassification as our criterion, which means that our attack is successful if the returned label with the highest probability differs from the correct label.

Evaluating the quality of adversarial images in detection is a challenge since a detector is largely different from a classifier and the quality
of adversarial examples cannot be properly measured based on the number of changed pixels only.
For attacking a classifier, the objective is to perturb as few pixels as possible.
For attacking detectors, however, people can still easily recognize the politician in a political image, even with many pixels have been perturbed.
If the attacker performs some political activities, such as insulting slogans, on this disturbing image, the detection service may fail to block such misdeed.
Instead, we consider three evaluation methods including $L_0$, Peak Signal to Noise Ratio (PSNR) and Structural Similarity (SSIM).
$L_0$ distance corresponds to the number of pixels that have been altered in an image.
We assume the original input is O, and the adversarial example is ADV. Then,

\begin{equation}
L_0: ||\mathbf{z}||_{0} = \#\{i|z_i \neq 0\}
\end{equation}
where $\mathbf{z}$=O-ADV.

We also use PSNR \cite{amer2002reliable} to measure the quality of images.
PSNR value is measured in dB. Typical PSNR values for visually-proper images are usually between 20 and 40 dB, where
higher is better \cite{amer2002reliable}.
\begin{equation}
PSNR = 10log_{10} (MAX^2/MSE)
\end{equation}
where $MAX =255$, and $MSE$ is the mean square error.
For an RGB image ($m\times n \times3$), 
\begin{equation}
MSE = \frac{1}{m\times n\times3}\times\sum_{b=0}^2\sum_{i=1}^n\sum_{j=1}^m ||ADV(i,j,b)-O(i,j,b)||^2
\end{equation}
where RGB means three-channel values of pixels, namely, red, green and blue, and $(i,j,b)$ is a coordinate of an image for channel $b$ ($0\leq b\leq2$) at location $(i,j)$.

To measure the image similarity, the SSIM index
is adopted in this paper \cite{wang2004image}.
Therefore, $L_0$ distance is used to measure how many pixels have been changed, while PSNR is used to measure the image quality, and SSIM measures the structural similarity.
In the following sections, the $L_0$ distance is used to present the $L_0$ distance between the original image and the adversarial image for short.

\section {Black-box Attack Algorithms}
In this section, we mainly discuss the black-box attacks used in our experiments.
We list five frequently-used image processing techniques, which can make image adversarial in Section \ref{41}.
In Sections \ref{42}-\ref{44}, we analyze the flaws of previous literature on adversarial example attacks and propose our attacks.
\begin{figure*}[!htbp]
	\centering
	\begin{tabular}{cccccccc}
	
		\includegraphics[width=0.14\linewidth]{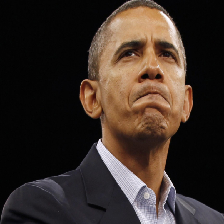}  &
		\includegraphics[width=0.14\linewidth]{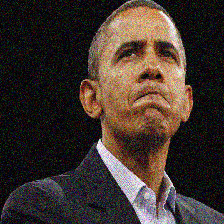}  &
		\includegraphics[width=0.14\linewidth]{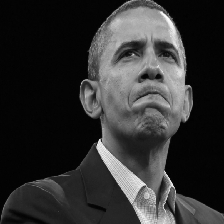}  &
		\includegraphics[width=0.14\linewidth]{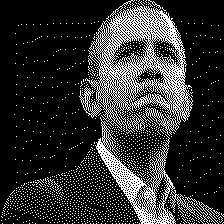} &
		\includegraphics[width=0.14\linewidth]{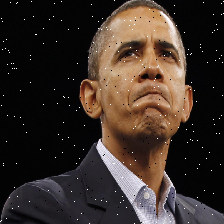}&
		\includegraphics[width=0.14\linewidth]{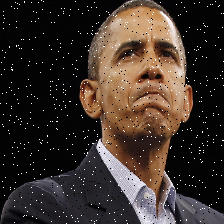}  \\
		(a) Origin & (b) Gaussian noise & (c) Grayscale & (d) Binarization &(e) $p$=0.05   &(f) $p$=0.15 \\
		\includegraphics[width=0.14\linewidth]{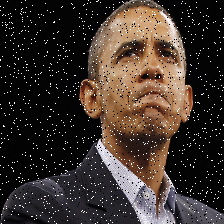}  &
		\includegraphics[width=0.14\linewidth]{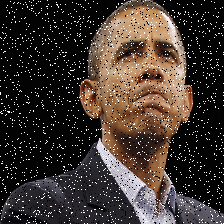}&
		\includegraphics[width=0.14\linewidth]{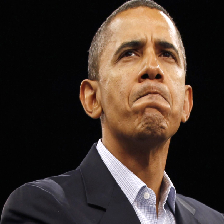}  &
		\includegraphics[width=0.14\linewidth]{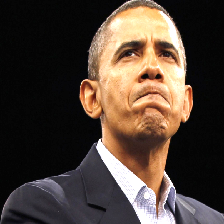}&
		\includegraphics[width=0.14\linewidth]{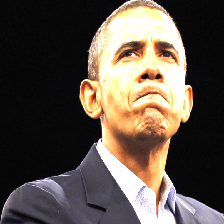}  &
		\includegraphics[width=0.14\linewidth]{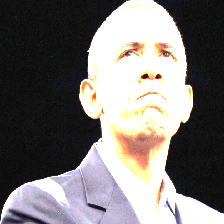}  \\

		(g) $p$=0.3 & (h) $p$=0.5& (i) $\varepsilon$=0.1& (j) $\varepsilon$=0.3 & (k) $\varepsilon$=0.5&(l) $\varepsilon$=0.8  \\
		
		

	\end{tabular}
	\caption{Five image processing techniques on a political image.
	According to the visibility of the image, we set the $\varepsilon$ between 0.1 and 0.8, and set $p$ between 0.05 and 0.5.
	(e)-(h) are Salt-and-Pepper noises with different parameters.
	(i)-(l) are Brightness Control attacks with different parameters.}\label{politicalexample}
	\vspace{-0.5em}
\end{figure*}
\subsection {Image Processing based Attack}\label{41}
In the following, we explore the effect of five different image processing techniques, including Gaussian Noise, Grayscale Image, Image Binarization, Salt-and-Pepper Noise, and Brightness Control, under which we take the images after the above image processing techniques as adversarial images directly.
The reason we choose these five techniques is that they are sufficiently representative.
As an empirical study, we hope our work could be easily extended to other image processing techniques. All these image processing techniques are implemented with Python libraries, such as \emph{skimage}\footnote{https://scikit-image.org}, \emph{OpenCV}\footnote{https://opencv.org}, and \emph{PIL}\footnote{http://www.pythonware.com/products/pil}, and we employ the parameters that can main well image visibility in our evaluation.

\subsubsection{\textbf{Gaussian Noise}}
Gaussian noise is statistical noise with a probability density function (PDF) equal to the normal
distribution as shown below.

\begin{equation}
f(x) = \frac{1}{\sqrt{(2\pi)}\sigma}exp(-\frac{(x-\mu)^2}{2\sigma^2})
\end{equation}
where $\sigma$ and $\mu$ represent the mean and variance, respectively.
Thus
\begin{equation}
ADV =O + Noise
\end{equation}
Note that, clipping \emph{ADV} is necessary to maintain a reasonable RGB value, which is [0, 255].
An example is shown in Fig. \ref{politicalexample} (b).

\subsubsection{\textbf{Grayscale Image}}
In a grayscale image, the value of each pixel is a single sample representing the amount of light, i.e.,
 it only carries intensity information.
 In the computer vision field, the black and white image contains only black and white pixels,
while the grayscale image has many levels of color depth between black and white.
To obtain the \emph{ADV}, the following equation\footnote{https://en.wikipedia.org/wiki/Grayscale} is used:
\begin{equation}
ADV = R*0.299 + G*0.587 + B*0.114
\end{equation}
This function is also implemented in Python library (\emph{PIL}).
Clipping the \emph{ADV} is necessary to maintain a reasonable RGB value.
An example is shown in Fig. \ref{politicalexample} (c).

\subsubsection{\textbf{Image Binarization}}
A binary image is a digital image that has two possible values for each pixel.
Typically, the two colors used for a binary image are black and white.
\emph{Floyd-Steinberg dithering} \cite{Floyd1975An} is used to approximate the original image luminosity levels in the implementation of \emph{PIL}.
An example is shown in Fig. \ref{politicalexample} (d).

\subsubsection{\textbf{Salt-and-Pepper Noise}}
Salt-and-Pepper noise is also known as impulse noise. This noise can result in sharp and sudden disturbances in the image signal.
For a pixel $(i,j,k)$ of an RGB image, the noise image of Salt-and-Pepper is calculated by $Noise(i,j,k;p)$,
where $Noise$ is the noise image,  and $p$ is the noise density. The other pixels remain original.
We show several examples in Fig. \ref{politicalexample} (e)-(h).

\subsubsection{\textbf{Brightness Control}}
We hypothesize that the image brightness may affect classification/detection results. Thus we iteratively adjust the brightness of an image to
observe the change of the result. A constant value is added to all the pixels of the image at the same time.
Then we clip it to a reasonable range, namely, [0, 255].

\begin{equation}
ADV = O +\varepsilon * 255
\end{equation}
where $\varepsilon$ is the parameter to control brightness, which is in the range of [0,1].
Several examples on brightness are shown in Fig. \ref{politicalexample} (i)-(l).


\subsection {Single-Pixel Attack}\label{42}
The single-pixel attack proposed by Nina et al.\cite{shiva2017simple}, is an attack where the perturbation of a single pixel would make the classifier generate a wrong label. However, it suffers from several limitations. First, when processing high-resolution images, a single pixel is usually not enough to cause the misclassification. Second, their experiment was completely offline and the data were fed to the classifiers directly. For the online classifier in the physical world, their attack is difficult to succeed.

To address these problems, we design a new single-pixel attack by gradually increasing the number of modified pixels and integrating the idea of image semantic segmentation.
In order to verify the validity of semantic segmentation, we implement the attack in three areas of an image, namely, subject region, non-subject region, and random region.
The random region is chosen as a baseline to compare with the other two regions.
These three regions are formally defined as follows:
\begin{itemize}
\item \emph {Subject region}: a region composed of all the pixels which belong to a \emph{subject class}.
\item \emph{Non-subject region}: a region composed of all the pixels which do not belong to any \emph{subject class}.
\item \emph{Random region}: a region composed of all the pixels chosen from the image randomly.
\end{itemize}

In the rest of this paper, we name the Single-Pixel attack as the SP attack since it is inspired by the single-pixel perturbation.

\subsection {Subject-based Local-Search Attack }\label{43}
Nina et al. \cite{shiva2017simple} also proposed a local greedy algorithm that searches the optimal perturbation of a local area iteratively. 
The optimal perturbation is defined as the one that has the largest influence on the model decision in each iteration.
However, there are flaws that make this attack ineffective on cloud classifiers or detectors.
Firstly, the image data is fed to a classifier directly. As indicated by Kurakin et al. \cite{kurakin2016adversarial}, the transformations applied to images in the process of printing them may
have a negative effect on adversarial examples.
Our experiment also shows that the JPEG format hinders adversarial examples to a certain extent, while the PNG format does not.
The reason is that the PNG format uses lossless compression coding. If the method in \cite{shiva2017simple} sent the perturbed RGB values to the VGG model \cite{chatfield2014return} and generated the adversarial example, it may be not adversarial
to online cloud-based classifiers anymore using the JPEG format.
Secondly, their method initialized a very large perturbation (RGB values exceed 500) to detect the probability changes of the classifier, which is impractical for the cloud-based classifier that accepts images with RGB values between 0 and 255.
Finally, if the initial disturbance region is very large, it is easy to fall into a local optimum, making it difficult to find adversarial examples.
Based on the above reasons, the local greedy algorithm \cite{shiva2017simple} is difficult to apply to cloud-based classifiers or detectors.

In this paper, we propose a Subject-based Local-Search (SBLS) attack by incorporating semantic segmentation to speed up the attack and saving all the images in the PNG format to retain their original features. Considering the online models, the initial modified pixel values are 0 or 255.
The main steps of our algorithm, as shown in Algorithm 1, are summarized as follows.
\begin{enumerate}
\item Firstly, we obtain the subject region of the image by semantic segmentation techniques.
\item Secondly, 50 pixels are selected from the subject region randomly and we perturb the image on each of the 50 pixels to generate 50 perturbed images.
The perturbed images are fed to the cloud-based detector and 50 predictions are produced.
The first $N$ (10 in this paper) pixels for which the probability drops the most are picked.
\item Thirdly, the image is perturbed on the $N$ pixels and the algorithm will record whether the prediction result changes from illegal to normal.
\item Finally, the perturbed image is taken as the initial image of the next round and cycle through the above steps until the label becomes normal or the generation fails.
\end{enumerate}

\begin{algorithm}[htb]
  \caption{ Subject-based Local Search (SBLS) Attack.}
  \label{alg:Framwork}
  \begin{algorithmic}[1]
    \Require   image: O,  local distance: D,
round: R,
 perturbation coefficient: P,
 the number of modified pixels per round: N,
 50 detections per cycle
 \Ensure  adversarial example or failure
\State S = subject(O)
\State r = 0
\While{$r<R$}
\State Locations = random(S, 50)
\For {axis in Locations}
\State		Imagetemp = perturb(O, axis, P)
\State		Label, Prob = cloud.predict(Imagetemp)
\If { Label ==normal}
			\Return Imagetemp
\EndIf
\State		Probs.append(Prob)
\EndFor
\State	Index = argsort(Probs)[:N]
\For{ i in Index}
\State		O = perturb(O,Locations[i],P)
\EndFor
\State	Label, Prob = cloud.predict(O)
\If { Label ==normal}
			\Return O
\EndIf
\State	Locations = Locations+D
\EndWhile\\
\Return Failure;
\end{algorithmic}
\end{algorithm}

In Algorithm 1, \emph{subject(O)} means getting the subject region of $O$, \emph{random(S, 50)} means getting 50 pixels of $S$ randomly, \emph{perturb(O, axis, P)} means the perturbation of $O$ on location $axis$ with
the coefficient $P$, and \emph{cloud.predict(Imagetemp)} means getting the label and probability of \emph{Imagetemp} from a cloud API.
Here, we assume that the cloud APIs will return both the label and the probability (score).
Since the initial images contain illegal contents, we iterate until the labels become normal or the algorithm fails.

\subsection {Subject-based Boundary Attack}\label{44}
Boundary Attack solely relies on the final model decision \cite{brendel2017decision}, which is also called the decision-based attack.
A decision-based attack starts from a large adversarial
perturbation and then seeks to reduce the perturbation while staying adversarial.
This method works in theory but not efficiently.
As reported, about 1.2 million predictions were used in the Boundary Attack to find an adversarial image for ResNet-50 \cite{brendel2017decision}, which is a huge overhead for the cloud service APIs.
If we want to generate hundreds of adversarial images, the required time and expense will be unbearable.

To make the attack practical, we design a new Subject-based Boundary (SBB) attack by incorporating the semantic segmentation and a greedy algorithm.
By semantic segmentation, the subject region is first perturbed with the average RGB value of the non-subject region, since the background color has a great influence on the recognition of a \emph{subject class} based on the previous experiments.
Through the greedy algorithm, the attack is able to recover as many of the perturbed pixels as possible, making the probability of correct classification as small as possible.
The main steps of the algorithm, as shown in Algorithm 2, are summarized as follows.
\begin{enumerate}
\item First, all pixels in the subject region are perturbed, which keep the image free of illegal content and the probability of being predicted to be illegal closing to zero.
\item Then, the ${L_0}$ distance between the current perturbed image and the original image is computed.
A certain percentage of pixels, which are selected from the different pixels between the original image and the perturbed one, are recovered randomly.
\item Next, the recovery process is repeated to choose the best recovering, which leads to the slowest increase in prediction probability or score, given that the perturbed image is still recognized as normal by the cloud API.
\item Finally, repeat steps 2-3 until the image is correctly classified.
If the perturbed image is recognized as illegal, the iteration will be stopped and the last perturbed image will be returned as the adversarial image.
\end{enumerate}

\begin{algorithm}[htb]
  \caption{ Subject-based Boundary (SBB) Attack.}
  \label{alg:Framwork}
  \begin{algorithmic}[1]
    \Require
      Oringin image: O,
	round: R,
	detetions: D,
    \Ensure  adversarial examples;

    \State S = subject(O),
    \State Non-S = O - subject(O),
    \State averpixel = getaverpixel(Non-S),
    \State ADV = perturb(O, S, Averpixel)
\While{$r<R$}
\State Step = $L_0$(O, adv)/10+100
\For	{$j=0$; $j<detections$; $j++$} :
\State		Advtemp = recover(step, O, ADV)
\State		Advcandidate.append(advtemp)
\State		Label, Prob = cloud.predict(Advtemp)
\If {Label ==normal}
\State	Probs.append(Prob)
 \EndIf		
 \EndFor
 \If {len(Probs)==0}
 \State break
 \EndIf
\State	Index = argmax(probs)
\State	ADV = Advcandidate[Index]
\EndWhile \\
\Return $ADV$;
  \end{algorithmic}
\end{algorithm}
In  Algorithm2, \emph{getaverpixel (Non-S)} means getting the average pixel value of a Non-S region,  \emph{$L_0$(O,ADV)} means getting the $L_0$ norm distance between $O$ and $ADV$,
 and \emph{recover(step, O, ADV)} means recovering \emph{step} pixels according to the difference of $O$ and $ADV$.

\section{Evaluation}

\subsection{Validation of Semantic Segmentation}
Since cloud-based detectors are based on classifiers and attacking detectors is more difficult than classifiers,
we first conduct the SP attack on classifiers to better understand the validity of semantic segmentation.
The results on classifiers can help us adjust our attack algorithms.
We choose the SP attack because it leverages coarse-grained perturbation.
If the SP attack works well with semantic segmentation techniques, SBLS and SBB attacks should have better performance since both of them leverage fine-grained perturbation.
Several local models and a cloud-based classifier are used in the experiment.
The local models we are using are VGG16 \cite{simonyan2014very}, Resnet50 \cite{he2016deep} and InceptionV3 \cite{szegedy2016rethinking}.
In this paper, we leverage the \emph{Keras} framework and pre-trained deep learning models\footnote{https://github.com/fchollet/deep-learning-models/releases} to conduct experiments.
These pre-trained models are trained with ImageNet since
it is a widely used standard dataset.
Further, due to the easier usage of Baidu APIs, we choose Baidu animal classifier as an example case in our evaluation.

\subsubsection {Datasets}

We prepare the dataset by selecting 100 animal images from the ImageNet dataset.
Because VGG16 and Resnet50 both accept input images of size $224\times224\times3$,
every input image is clipped to the size of $224\times224\times3$.
Here, we select animal images for simplicity.
The attack strategy mainly consists of two parts: perturbation methods and perturbation regions.

\begin{figure}[htbp]
	\centering
	\begin{tabular}{ccc}
		\includegraphics[width=0.28\linewidth]{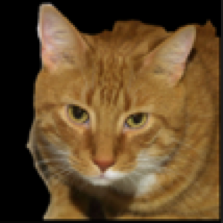}  &
		\includegraphics[width=0.28\linewidth]{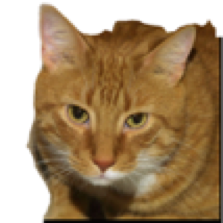}  &
		\includegraphics[width=0.28\linewidth]{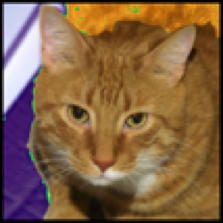} \\
		(a) $P$=0 & (b) $P$=255 & (c) $P$=2\\
	\end{tabular}
	\caption{The pixels of background are set as different $P$ values. Here we just change the pixels of the non-subject region.} \label{catback}
	\vspace{-0.5em}
\end{figure}

\begin{itemize}
\item Different methods of perturbation have different effects on the prediction results.
Three types of perturbation are considered: $P$=0,
$P$=255, and $P$=2, where $P$ is a perturbation parameter.
For instance, $P$=0 or 255 means setting the RGB value of the pixel to 0 or 255, where 0 represents black and 255 represents white.
$P$=2 means multiplying the pixel value by 2 and clipping it to a reasonable range. An example of different perturbation methods is shown in Fig. \ref{catback}.

\item In order to verify the effectiveness of semantic segmentation in perturbation, the three perturbation regions described in Section 4.2 are chosen to test.
\end{itemize}

\subsubsection {Results and Analysis}
First, we carry out a precursor experiment by 1) perturbing pixels that do not belong to \emph{subject class} using different perturbation methods; 2) recording the changes in prediction.

The overall result is shown in Table \ref{misrate}.
From Table \ref{misrate}, it is evident that the perturbation of the non-subject region can cause a misclassification rate of up to 0.8.
VGG16 is the least resilient one, while InceptionV3 and Baidu are the two most robust models.
Besides, the results of $P$=2 have the lowest successful misclassification rate among the three perturbation methods since it is a slight perturbation.
We can also see from the result that about 60\% image classifications of InceptionV3 are not affected by the perturbation on the background pixels.
Thus, it is necessary to perturb the subject pixels to improve the power of the attack.

\begin{table}[!htbp]
\centering
\caption{Successful misclassification rate. We perturb all the pixels of the non-subject region and record the corresponding misclassification rate.}\label{misrate}
\begin{tabular}{ccccc}
\hline

\hline

\hline
Background & VGG16 & Resnet50 &InceptionV3 & Baidu (Online) \\
\hline

\hline

$P$=255 &  0.8 &0.58 &0.38 &0.40 \\
$P$=2 &  0.53&0.41 &0.25 &0\\
$P$=0  &0.68 & 0.56&0.39 &0.40 \\

\hline

\hline

\hline
\end{tabular}
\end{table}

\begin{figure*}[!htbp]
	\centering
	\begin{tabular}{ccc}
		\includegraphics [width=0.32\linewidth]{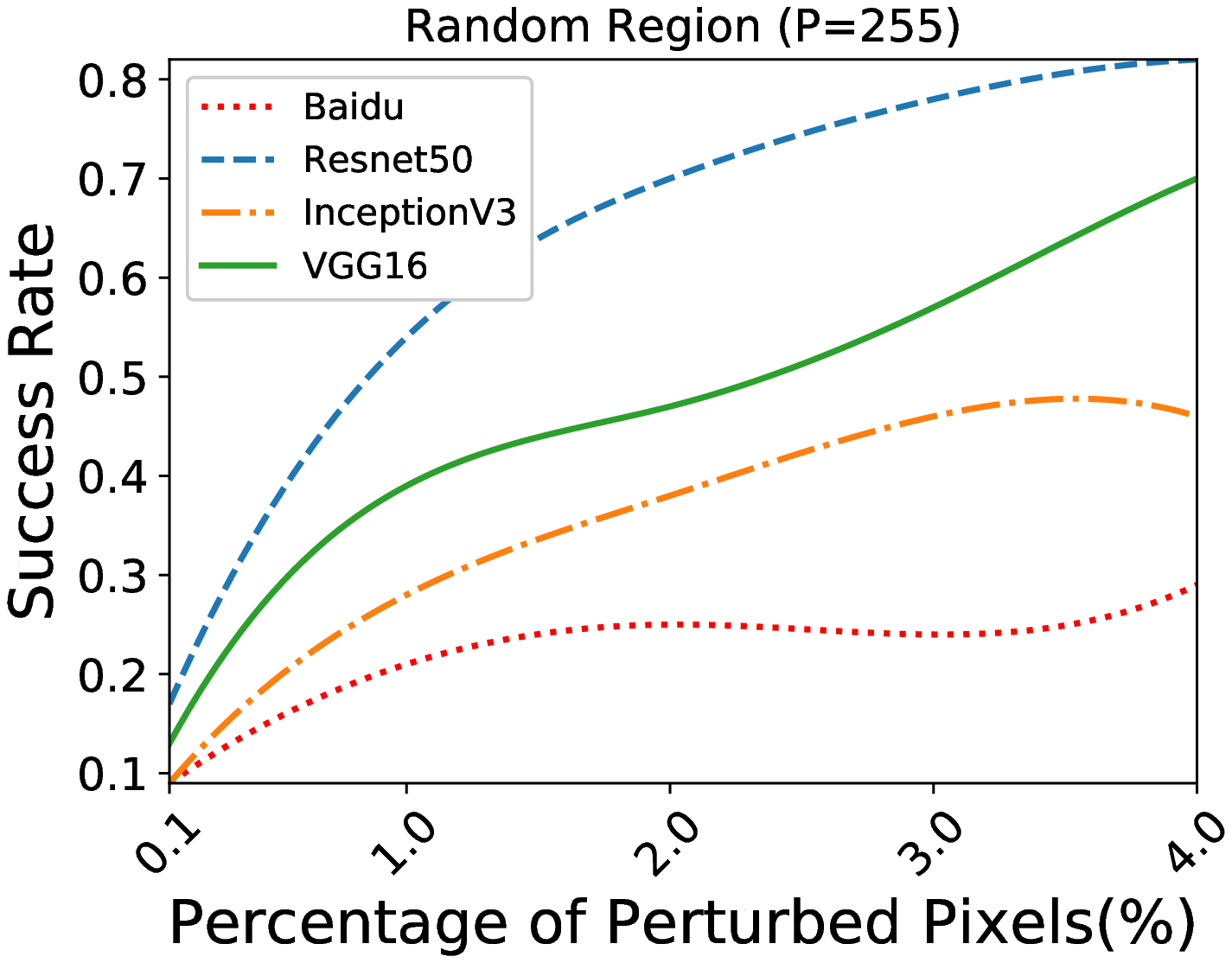}  &
		\includegraphics [width=0.32\linewidth]{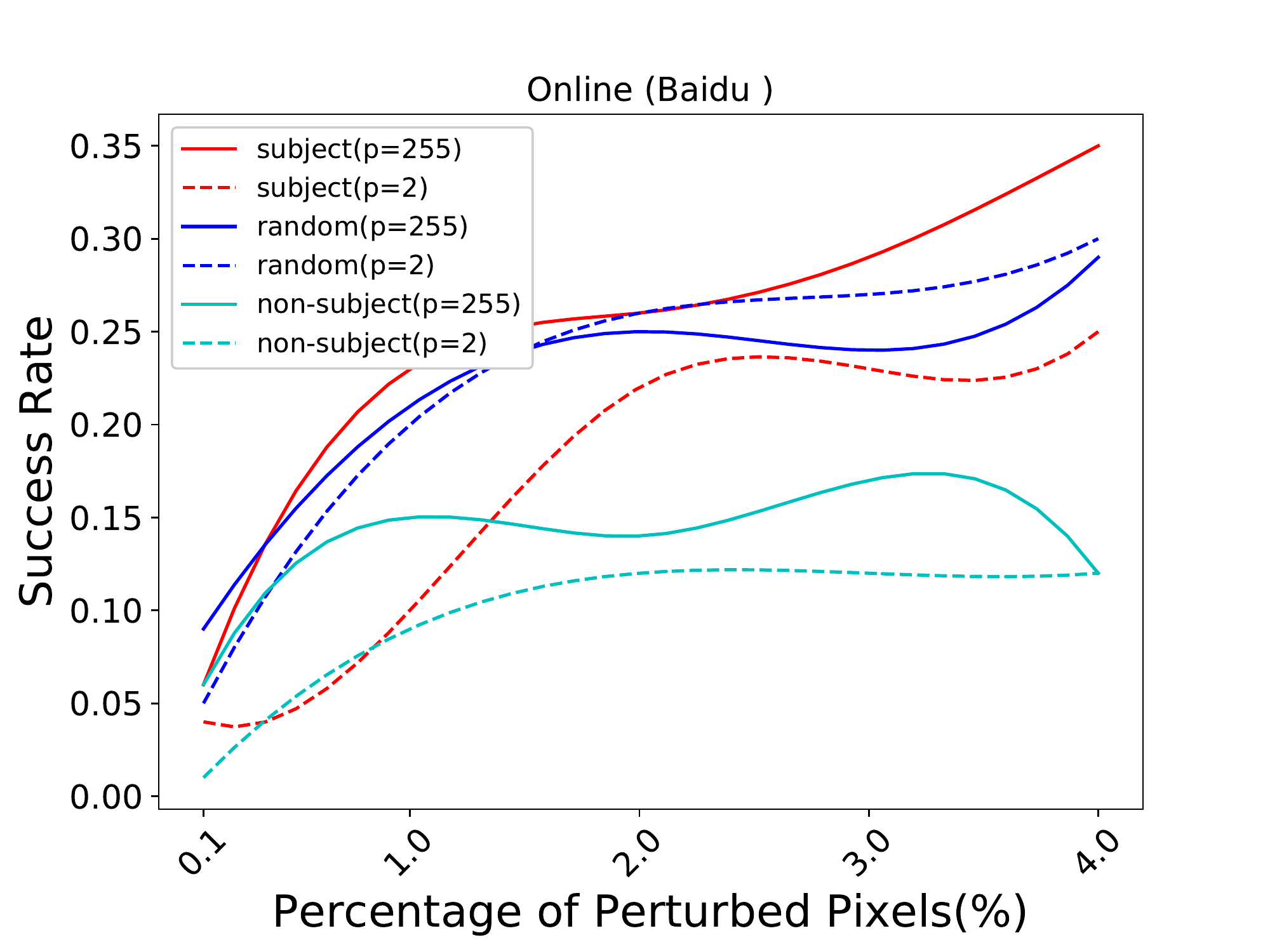}&
		\includegraphics [width=0.32\linewidth]{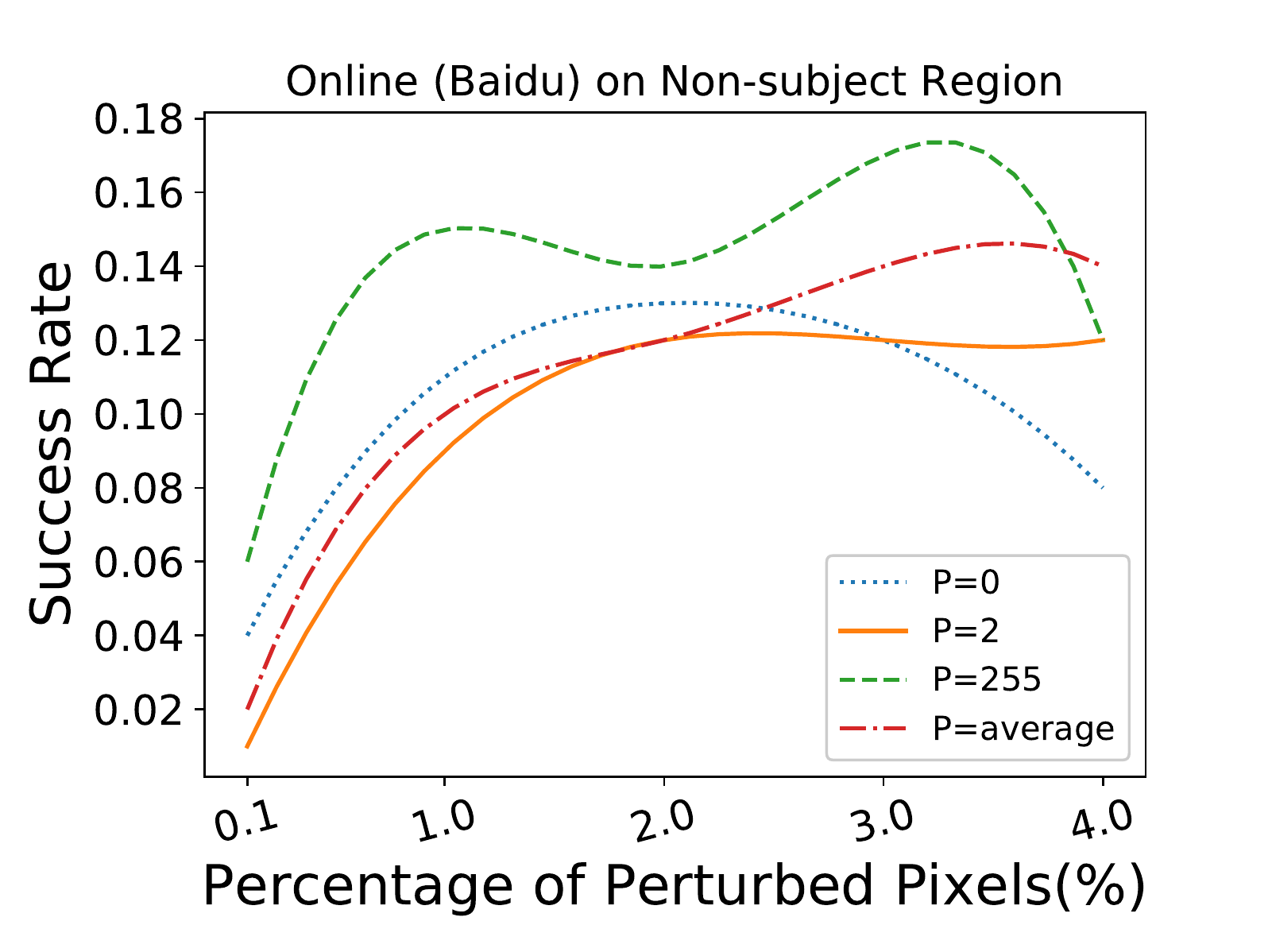}\\
		(a)&(b)&(c)\\
	\end{tabular}
	\caption{In (a), we randomly select pixels on the image to perturb and increase the number of perturbed pixels. The figure records the number of misclassifications under different numbers of perturbed pixels.
In (b), we test different regions of the image with $P=255$ and $P=2$ and record the number of successful misclassification. The horizontal axis is the number of perturbed pixels, and the vertical axis is the number of successful misclassification.
In (c), we perturb the pixels in the non-subject region with different $P$ values. $P= average$ means that we perturb the non-subject region with the average RGB value of the subject region.}\label{onlinebaidu}
\end{figure*}

We conduct a number of experiments to understand the choice of perturbation parameters and regions on different models, as shown in Fig. \ref{onlinebaidu}. Firstly, we adjust the number of perturbed pixels and record the effects on four classifiers.
As shown in Fig. \ref{onlinebaidu} (a), the number of successful attacks increases as the number of perturbed pixels increases.
After reaching a large perturbation, the success rate increases very slowly.
Compared to the local classifiers, attacking Baidu is more difficult.

Secondly, for a single classifier, we select the perturbation regions in three different ways to observe the changes in the prediction results.
We take the online classifier Baidu for example.
Fig. \ref{onlinebaidu} (b) shows that it is very sensitive to perturb the pixels in the subject region, resulting in a high misclassification rate.
Fig. \ref{onlinebaidu} (b) also demonstrates the results when we use $P$=2 to perturb the pixels.
However, surprisingly, the perturbation in random regions performs even better than subject regions.
Our conjecture is that slight perturbation in the non-subject region makes the pixels after perturbation close to those in the subject region.
In order to verify our conjecture, we perturb the non-subject pixels with the average RGB value of the subject region.
For all pixels in the subject region, the average RGB value in each channel is computed and the non-subject region is perturbed with the derived average value.
The results in Fig. \ref{onlinebaidu} (c) demonstrate that the perturbation with the average value performs better than that with $P$=2.
When perturbing 2000 pixels, the perturbation with the average value performs the best.

\textbf{Conclusions}. We draw the following conclusions based on the results from classifiers:
\begin{itemize}
\item Among our evaluation, the online models of Baidu have \textbf{relatively good robustness} and attacking its online models is more difficult than attacking local models using the same number of perturbed pixels.
\item Perturbations in \textbf{subject regions} are more effective than other regions.
\item The prediction of the model is sensitive to the magnitude of the disturbance value, especially the value close to the pixels of the subject region.
\end{itemize}

\subsection{Attacking Cloud-based Detectors}\label{attackdetector}
\subsubsection {Datasets and Preprocessing}
To test cloud-based detectors, 300 images are selected from Google Images or Baidu Images.
For the three kinds of illegal images to be detected, 100 images for each kind are manually selected and labeled.
All the images are resized to a fixed size: $224\times224\times3$.
Among the detectors we tested, they accept most image formats, such as JPG, PNG, JPEG, BMP, etc.
It is worth to mention that for our real-life cloud-based setting, we have to adjust these images properly followed the requirements of the tested APIs, including input format, size, and resolution. All these images were collected legally.
In order to avoid image loss during image compression and transmission, all the images in our experiments are saved with the PNG format and sent directly to the cloud API interface without additional transformations.
Because the images are labeled by us, the illegal images may not be identified as illegal by the evaluated detectors.
Therefore, we first filter the not qualified images by calling the detectors.
Specifically, the detectors will return the corresponding predictions, which are made up of probabilities (scores) and labels. Then, we discard the images with ambiguous labels while keeping the images labelled with illegal by both us and the evaluated detectors. For instance, Alibaba offers an option for manual reviewing to suspicious images and we exclude such images.
For Google's detector, we can only obtain a single-world result, e.g., \emph{POSSIBLE}, \emph{LIKELY} as shown in Table \ref{forms}.
Only the images whose labels are \emph{LIKELY} and \emph{VERY LIKELY} are kept.
Finally, the predictions of the remaining images are recorded in Table \ref{correct}, where ``--" means the platform does not provide the API service.
\begin{table}[!htbp]
\centering
\caption{Correct label by cloud APIs.}\label{correct}
\begin{tabular}{ccccc}
\hline

\hline

\hline
Platforms  & Pornography & Violence &Politician \\
\hline

\hline

\hline
Baidu &95/100&32/100&45/100 \\
Google&90/100&30/100&--\\
Alibaba&67/100&67/100&49/100\\
Azure &54/100&--&--\\
AWS&84/100&--&57/100\\

\hline

\hline

\hline
\end{tabular}
\end{table}

\begin{figure*}[!htbp]
	\centering
	\begin{tabular}{cccc}
		\includegraphics [width=0.23\linewidth]{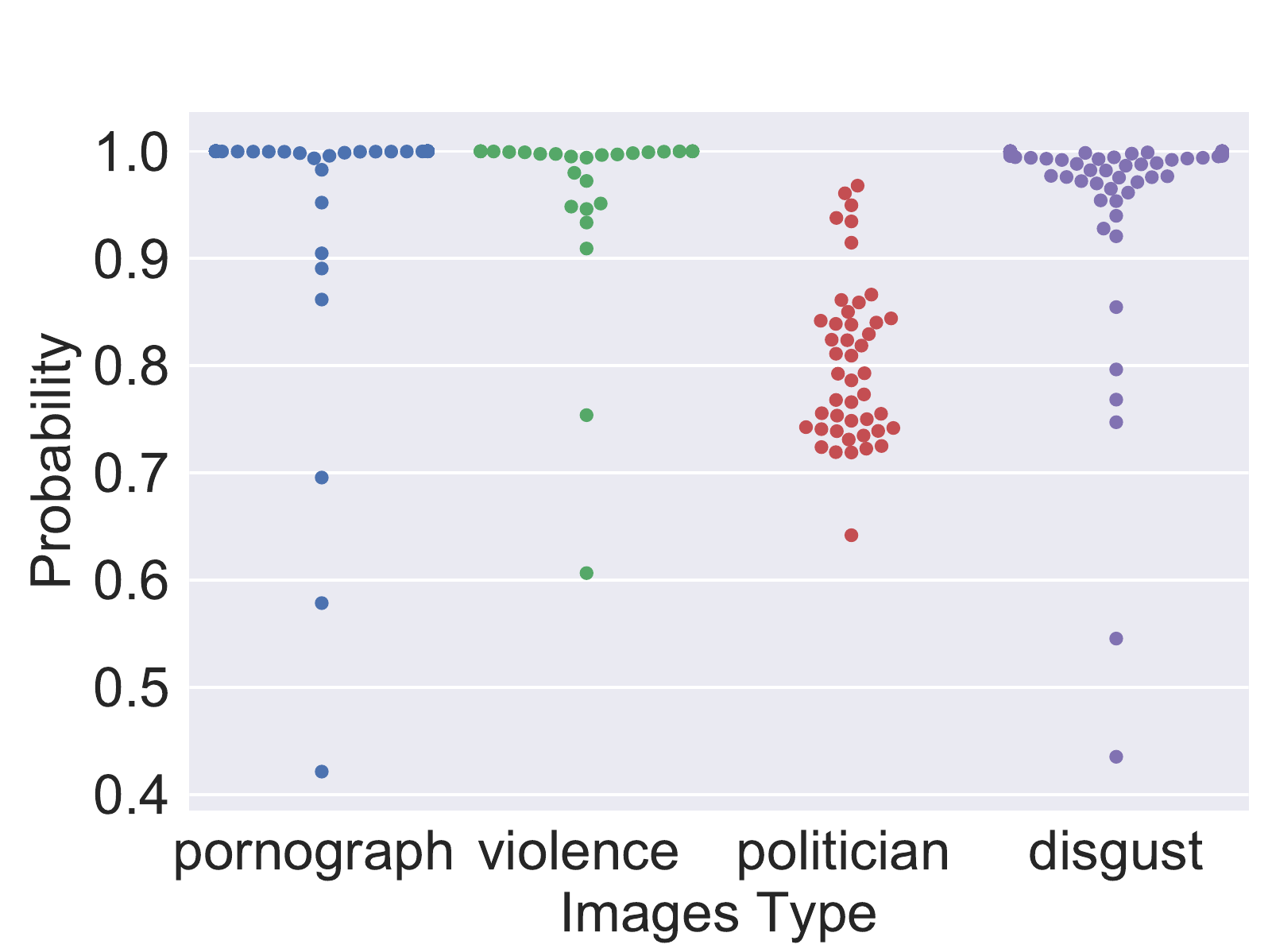} &
		\includegraphics [width=0.23\linewidth]{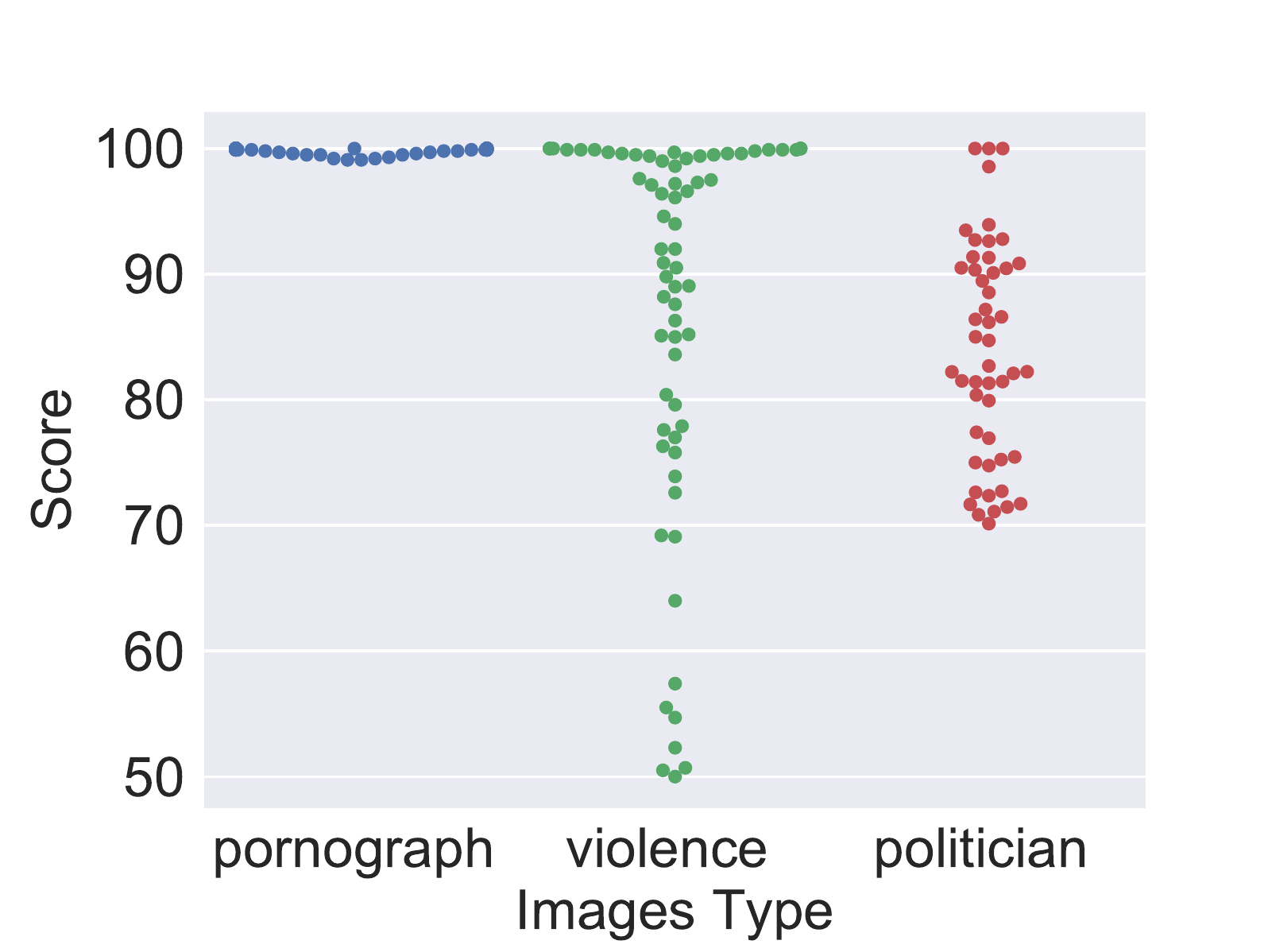}&
		\includegraphics [width=0.23\linewidth]{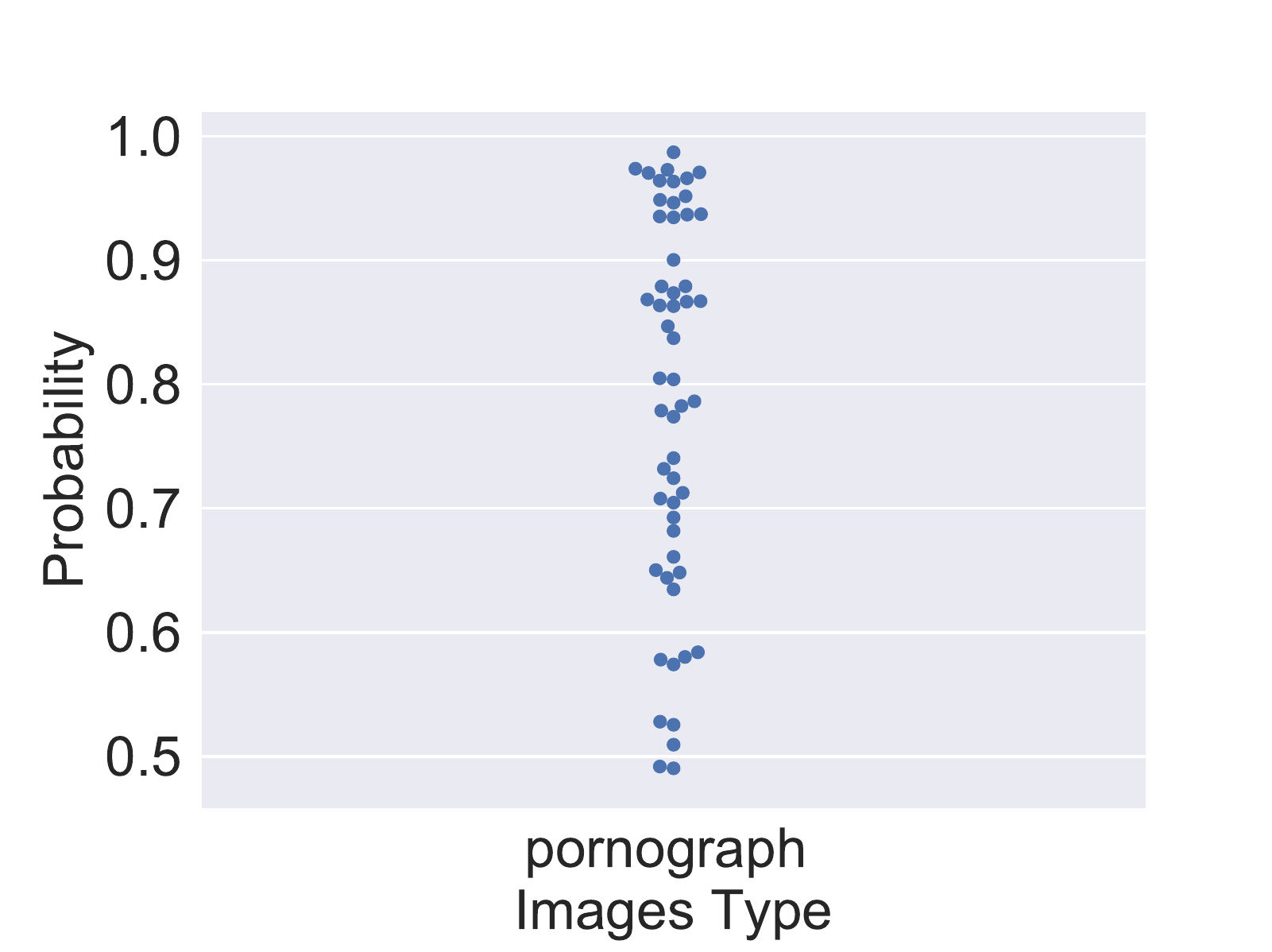}&
		\includegraphics [width=0.23\linewidth]{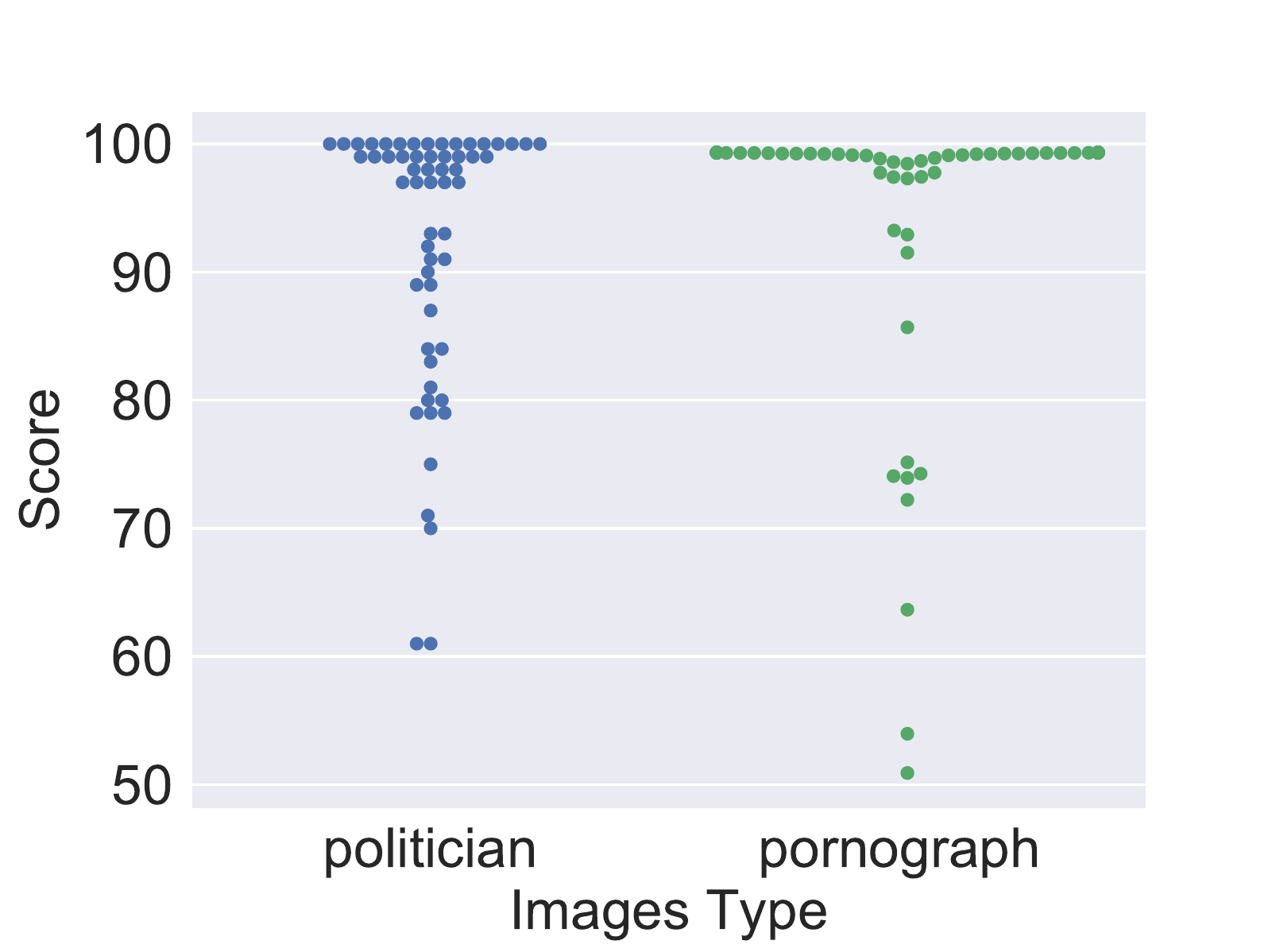}\\
		(a) Baidu&(b) Alibaba&(c) Azure&(d) AWS\\
	\end{tabular}
	\caption{The probabilities and scores of test images. Most points are concentrated around high probabilities.
	Besides, many points are overlapped where the probability is 1 or the score is 100.} \label{preprob}
\end{figure*}

According to Table \ref{correct}, for porn images, we can learn that Baidu, which labels 95\% of them correctly,
has done a better job than other cloud platforms.
To our surprise, 46\% of the pornographic images have not been identified by Azure's detector.
For violent images, Alibaba's detector has the best performance and it labels 67\% of them correctly. The detectors of Google and Baidu only recognize 30\% and 32\% of the violent images, respectively.
One possible reason is that the scenes of violent images are more complex and the detectors do not consider a variety of scenarios.

To better understand the quality of these images, we record the probabilities or scores when the API is called to detect these images.
The detailed information can be found in Fig. \ref{preprob}.
As shown in Fig. \ref{preprob},  the majority of images are labeled by the APIs with very high confidence.
For Azure's detector, 80\% of the probability labels are over 0.7.
It is more convincing for our evaluation to leverage images that are correctly classified by the detectors with high confidence. 
Otherwise, a small perturbation may cause the images to evade the detection with low confidence, which cannot indicate the strength of our attacks.

In our experiments, considering misclassification only is not enough.
The detector may give a similar prediction to illegal images though the predictions changed.
For example, the prediction may go from \emph{VERY LIKELY} to \emph{LIKELY}, which does not
make the detector give a completely opposite prediction.
Therefore, the successful adversarial examples in our experiments are defined as the examples that have completely changed the prediction.
Then, we use the three metrics in Section \ref{metrics} to measure the quality of these adversarial examples.

\subsubsection{Detectors of Pornographic Images}

The Internet is flooded with pornographic images, which is a serious problem for website regulators.
Websites often leverage detectors to detect these illegal images. Evasion attacks on these detectors can result in serious content security risks. Our considered cloud platforms all provide the pornographic image detection service as shown in Table \ref{forms}.

\begin{table}[!htbp]
\centering
\caption{Forms of prediction. We record the returned information and labels for every platform.}\label{forms}
\begin{tabular}{ccc}
\hline

\hline

\hline
Cloud & Feedback& Label Category\\
\hline

\hline

\hline
Baidu & Probability &Porn, Sexy, Normal\\
Google & N&  \tabincell{c}{UNKNOWN, VERY\_UNLIKELY, \\UNLIKELY, POSSIBLE, LIKELY, \\VERY\_LIKELY}\\
Alibaba & Score & porn, sexy, normal\\
Azure & Probability&True, False \\
AWS &Score & \tabincell{c}{Explicit Nudity, Nudity, \\Graphic Female/Male Nudity, \\Sexual Activity, Suggestive, \\Female/Male Swimwear Or Underwear,\\ Revealing Clothes, No label (normal)}\\

\hline

\hline

\hline
\end{tabular}
\end{table}

\begin{figure*}[!htbp]
	\centering
	\begin{tabular}{cc}
		\includegraphics [width=0.45\linewidth]{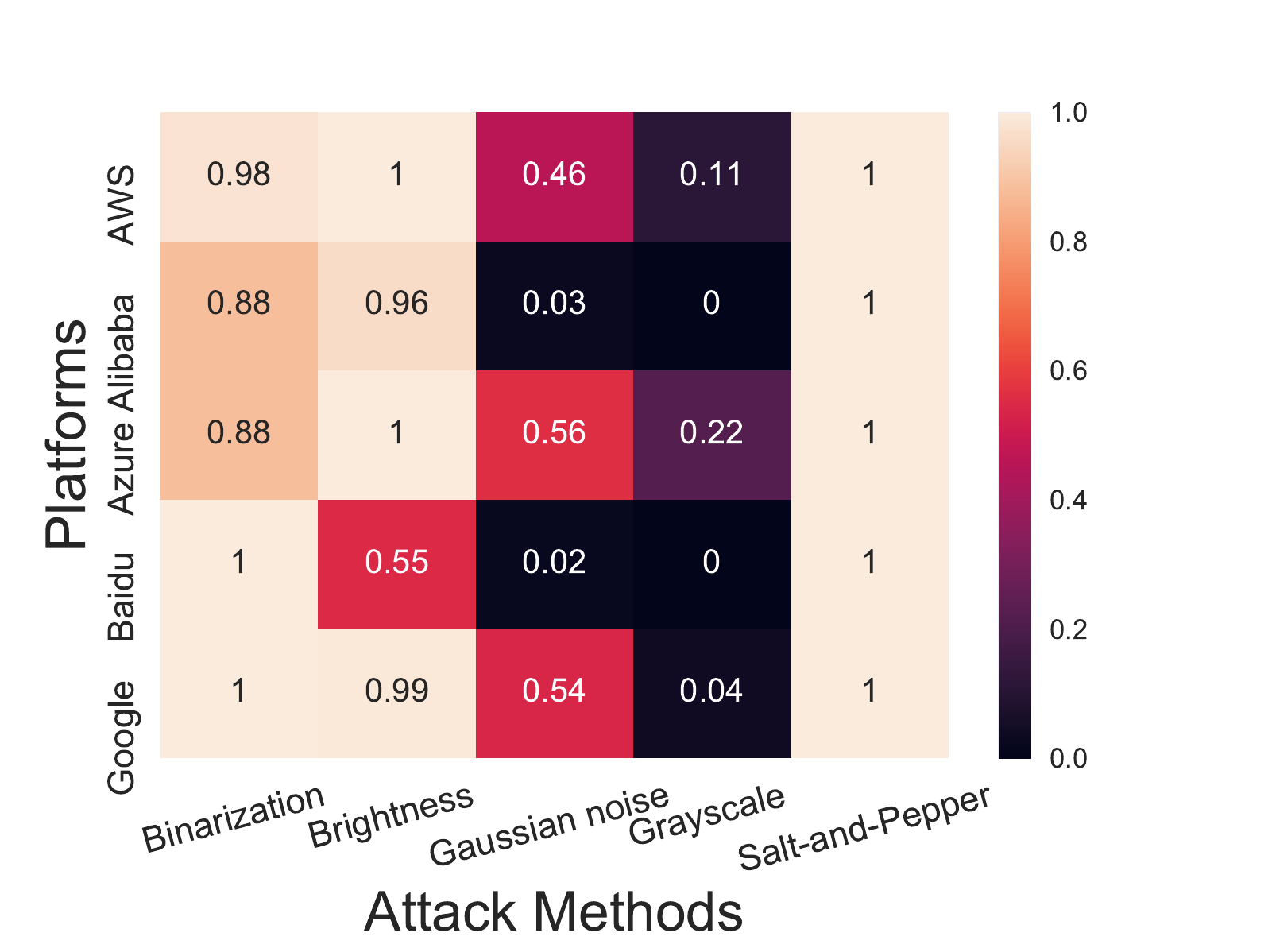} &
		\includegraphics [width=0.45\linewidth]{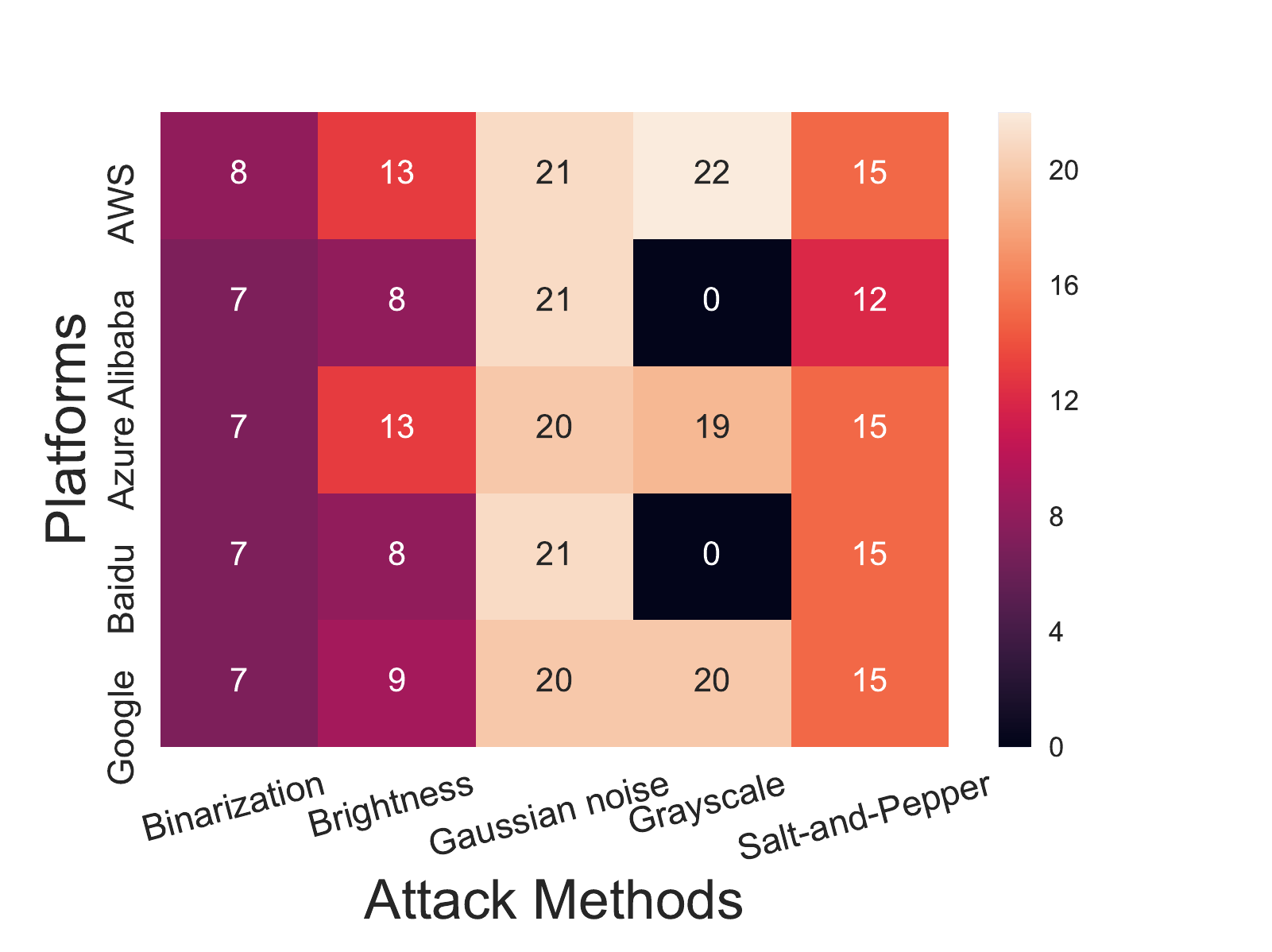}\\
		(a) Success rates&(b) PSNR values\\
	\end{tabular}
	\caption{The performance of the IP attack on pornographic images. } \label{sex-noise}
\end{figure*}

\begin{figure*}[!htbp]
	\centering
	\begin{tabular}{cc}
	
		\includegraphics[width=0.45\linewidth]{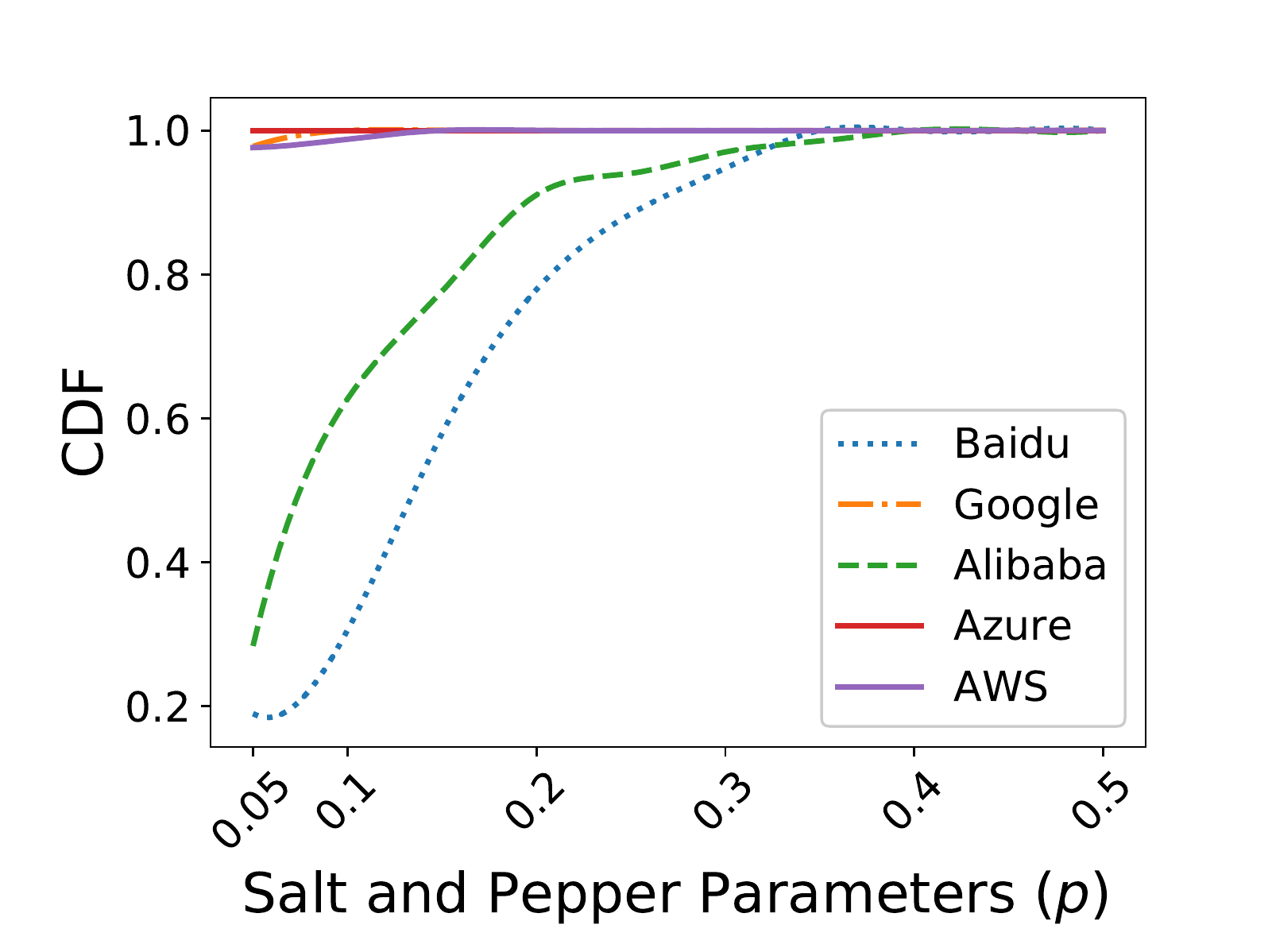}  &
		\includegraphics[width=0.45\linewidth]{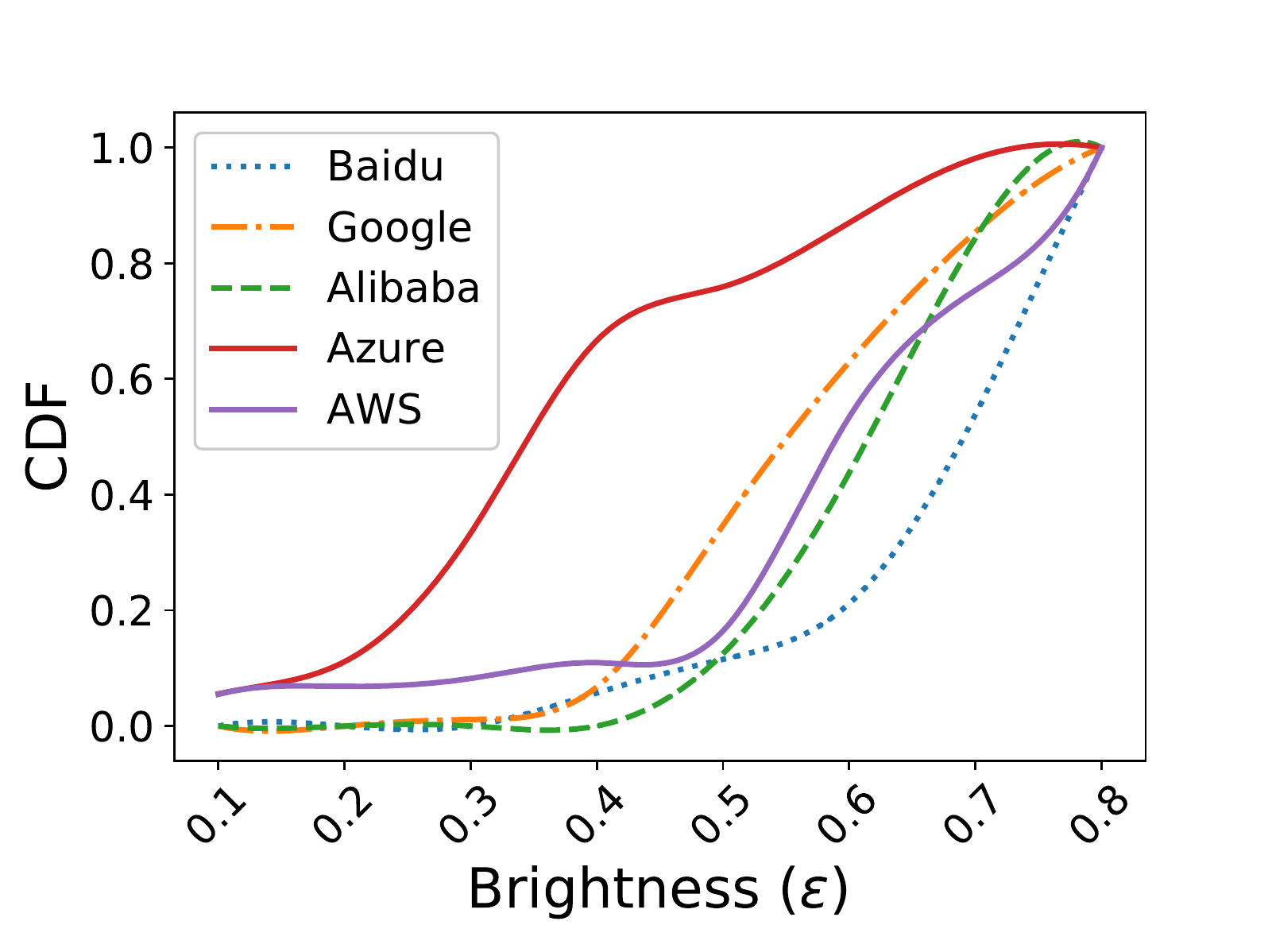} \\
		(a)&(b)\\

	\end{tabular}
	\caption{Cumulative distribution function. (a) is the CDF of the success rates for the Salt-and-Pepper attack and (b) is the CDF of the success rates for the Brightness attack.
	The horizontal axis is the corresponding parameter.}\label{CDF}
	\vspace{-0.5em}
\end{figure*}
{\textbf{Image Processing.}} The success rates of IP attacks are shown in Fig. \ref{sex-noise} (a).
From Fig. \ref{sex-noise} (a), we find that the detectors of Azure, AWS, and Google on pornographic images are vulnerable to the Gaussian noise attack.
The Grayscale attack has a slight effect on Google and Azure, and no effect on Baidu and Alibaba.
For Salt-and-Pepper and Brightness attacks, their success rates all increase as the parameters ($\varepsilon$ and $p$) increase.

In order to further evaluate the successful adversarial images, the PSNR value\cite{amer2002reliable} is used.
The average PSNR value of the successful adversarial images is shown in Fig. \ref{sex-noise} (b).
As shown in Fig. \ref{sex-noise} (b), we can learn that the results of Gaussian and Grayscale attacks are very promising.
The Binarization attack makes pornographic images contain only black and white pixels, which significantly reduces the content visibility.
Thus, the PSNR values are very small.
For Salt-and-Pepper and Brightness attacks, we keep increasing the attack parameters ($\varepsilon$ and $p$) until the attack is successful.
The CDF plots of the success rates are shown in Fig. \ref{CDF}.
In Fig. \ref{CDF} (a), we can see that the success rate reaches \textbf{100\%} when the $p$ of Salt-and-Pepper is \textbf{0.1} for the detectors of Google, AWS and Azure.
However, the detectors of Baidu and Alibaba require more Salt-and-Pepper noise to achieve high success rates.
In Fig. \ref{CDF} (b), we see that the Brightness attack needs a large $\varepsilon$ to achieve high success rates, which on the other hand affects the image quality greatly.

{\textbf{Single-Pixel Attack.}}
\begin{figure*}[htbp]
	\centering
	\begin{tabular}{cc}
		\includegraphics [width=0.45\linewidth]{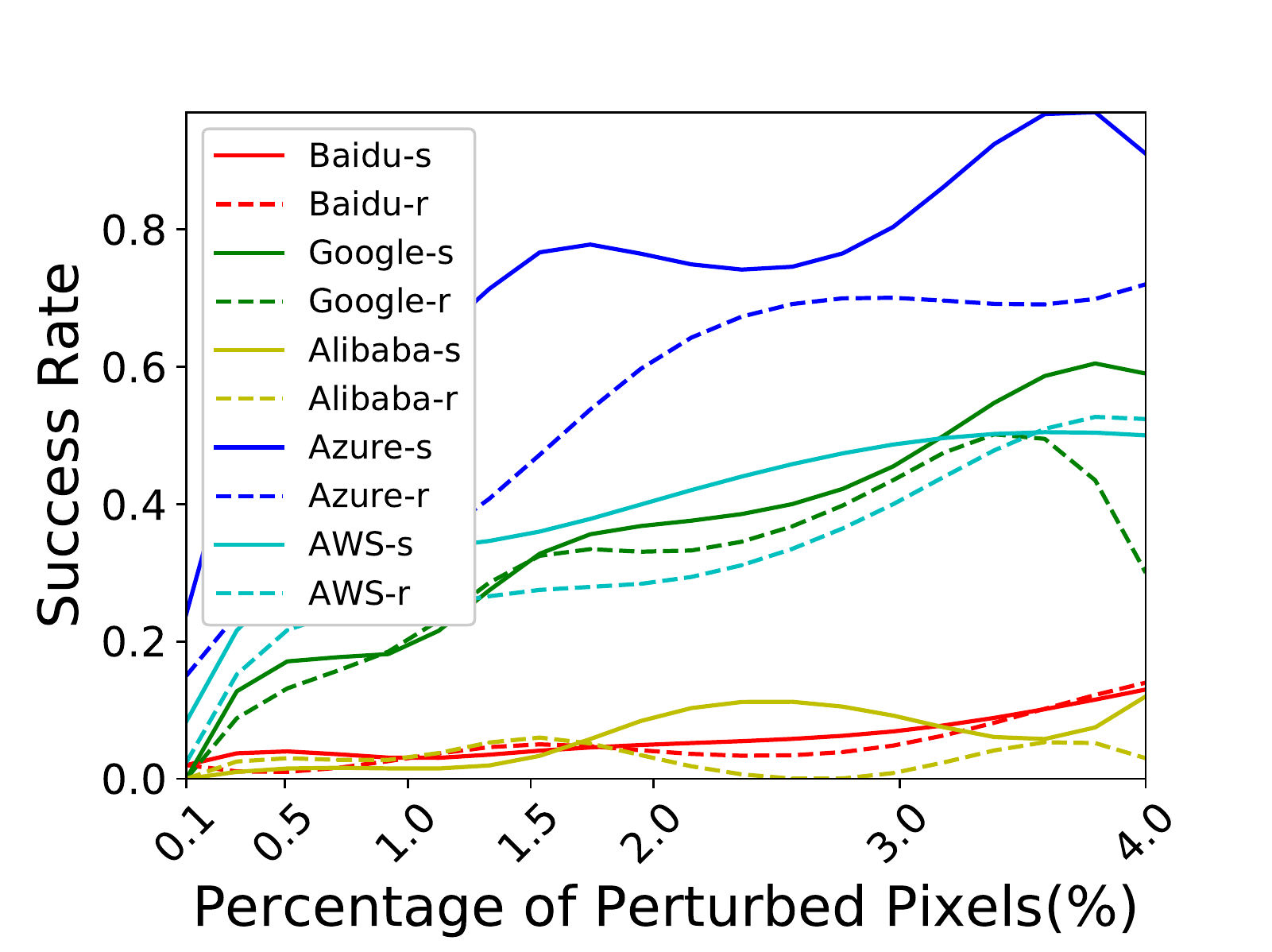}&
		\includegraphics[width=0.45\linewidth]{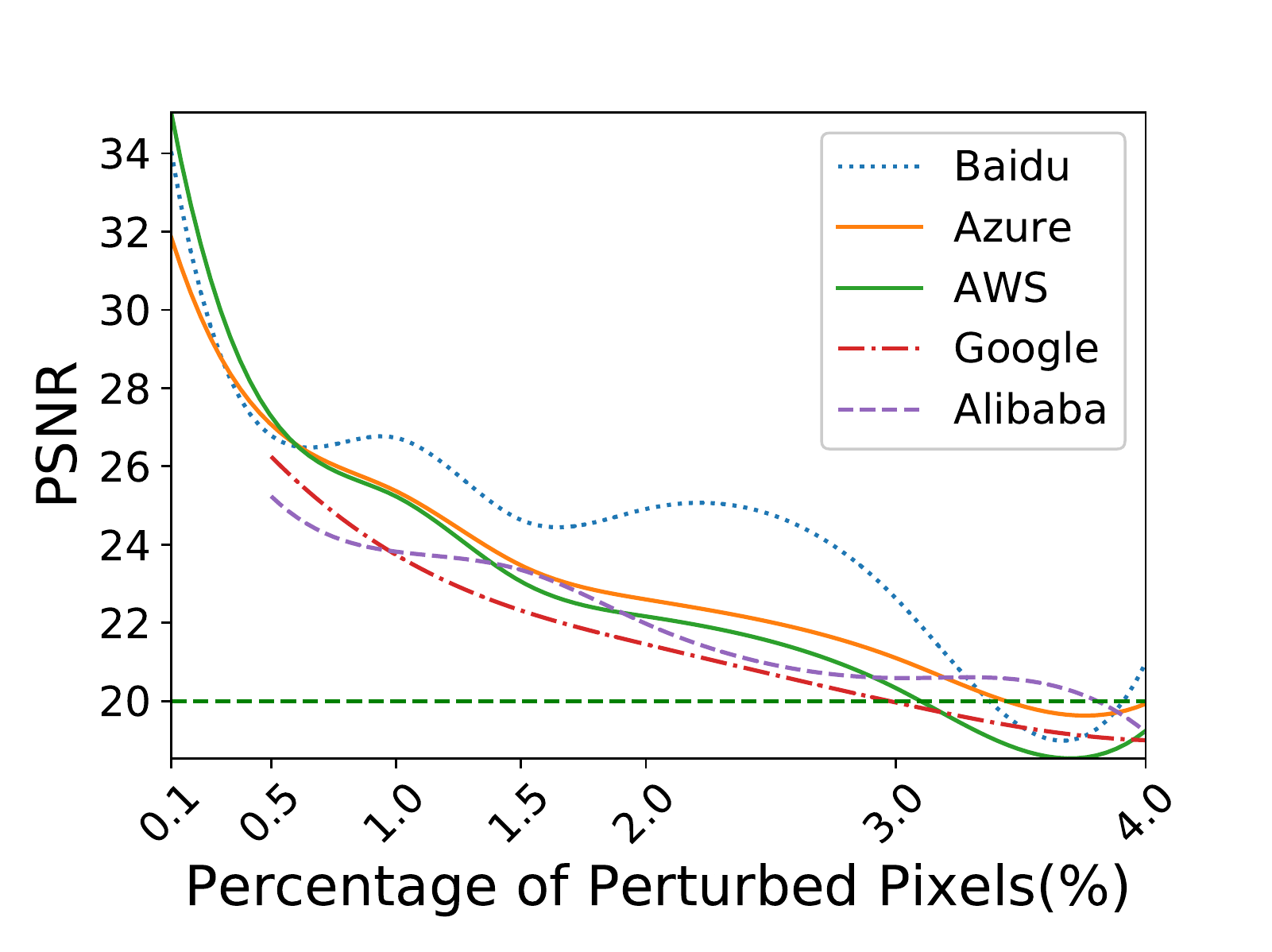}  \\
		 (a) Success Rates & (b) PSNR\\
	\end{tabular}
	\caption{(a) is the results of SP on pornographic images. The solid line represents the results of the subject region, and the dotted line represents the random region.
	(b) is the evaluation of SP on the subject region. The PSNR values of adversarial images are shown and 20 is the normal value of PSNR.}\label{singlesex}
	\vspace{-0.2em}
\end{figure*}
We conduct the Single-Pixel (SP) attack on the cloud platforms with perturbation in different regions of images, as shown in Fig. \ref{singlesex} (a),
where \emph{*-s} means the perturbation of the subject region,
and \emph{*-r} means the perturbation of the random region.
As shown in Fig. \ref{singlesex} (a), the effect of the perturbation in the \textbf{subject region} is much more significant than that in the random region,
which verifies the validity of semantic segmentation. Besides, we find that the attack success rate can be increased by perturbing more pixels on the subject regions.
Moreover, even if 2000 pixels are perturbed, the success rates of attacks on Baidu and Alibaba still remain low, whereas \textbf{91\%} of the same perturbed pornographic images can bypass
Azure's detector. This demonstrates that the detectors of Baidu and Alibaba are more resilient to the SP attack than others.

Now, we evaluate the image quality after perturbing the subject regions.
The PSNR values are shown in Fig. \ref{singlesex} (b).
The majority of the PSNR values are larger than 20, which means the quality of most successful adversarial images are in the acceptable range.

{ \textbf{Subject-based Local-Search Attack.}}
We set the maximum number of cycles $R$ to 30 in the Subject-based Local-Search (SBLS) attack.
Besides, we set $P$ to 255 and $D$ to 10, which correspond to the RGB value and expanding 10 unit pixels in the next loop as described in Algorithm 1, respectively.
Previous experiments have demonstrated the effectiveness of semantic segmentation.
Thus, the evaluation of subject-based adversarial images is statistically analyzed in the following subsections.
We skip Google's detector for the SBLS attack since Google's does not return probability or score that the SBLS attack is relying on. We also do not conduct the SBLS attack or SBB attack on AWS's detector since AWS only provides 5000 queries each month for free, which is not enough for us to test hundreds of images with the SBLS attack or SBB attack.
\begin{figure}[!htbp]
	\centering
	\begin{tabular}{ccc}
		\includegraphics [width=0.3\linewidth]{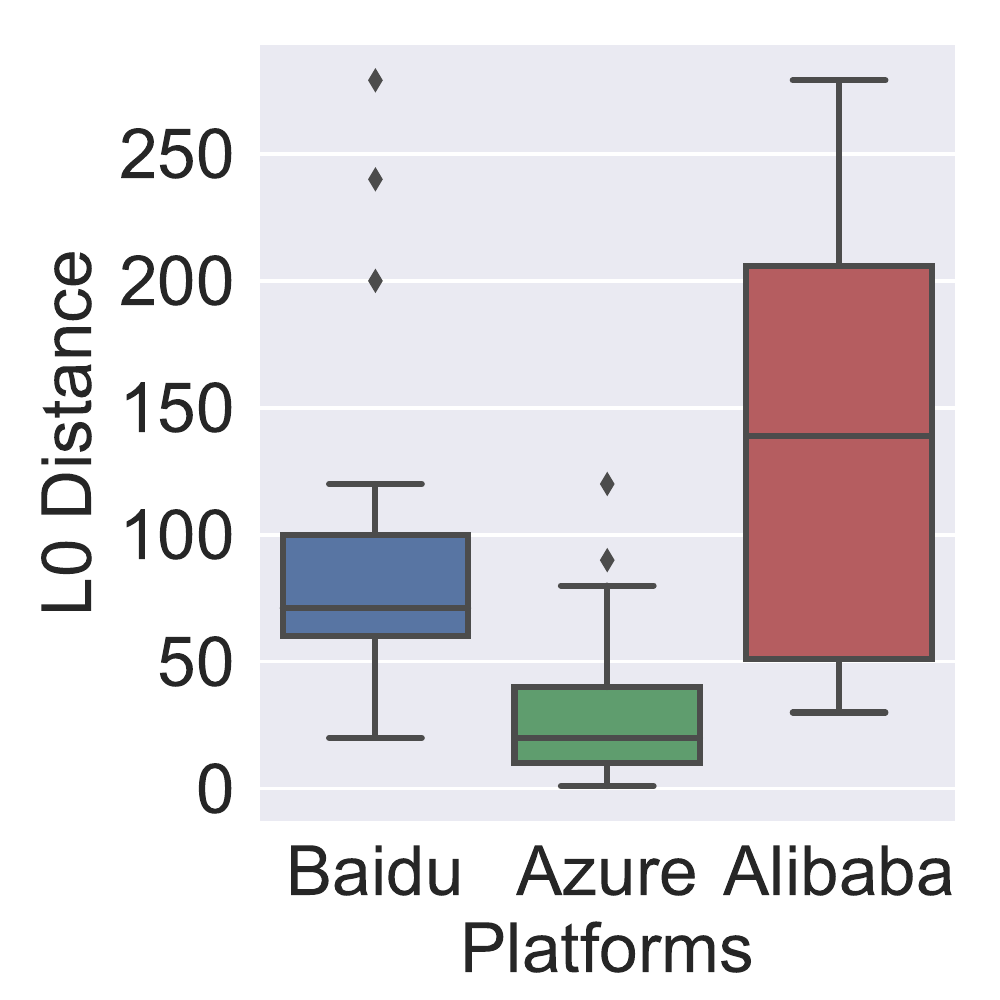} &
		\includegraphics [width=0.3\linewidth]{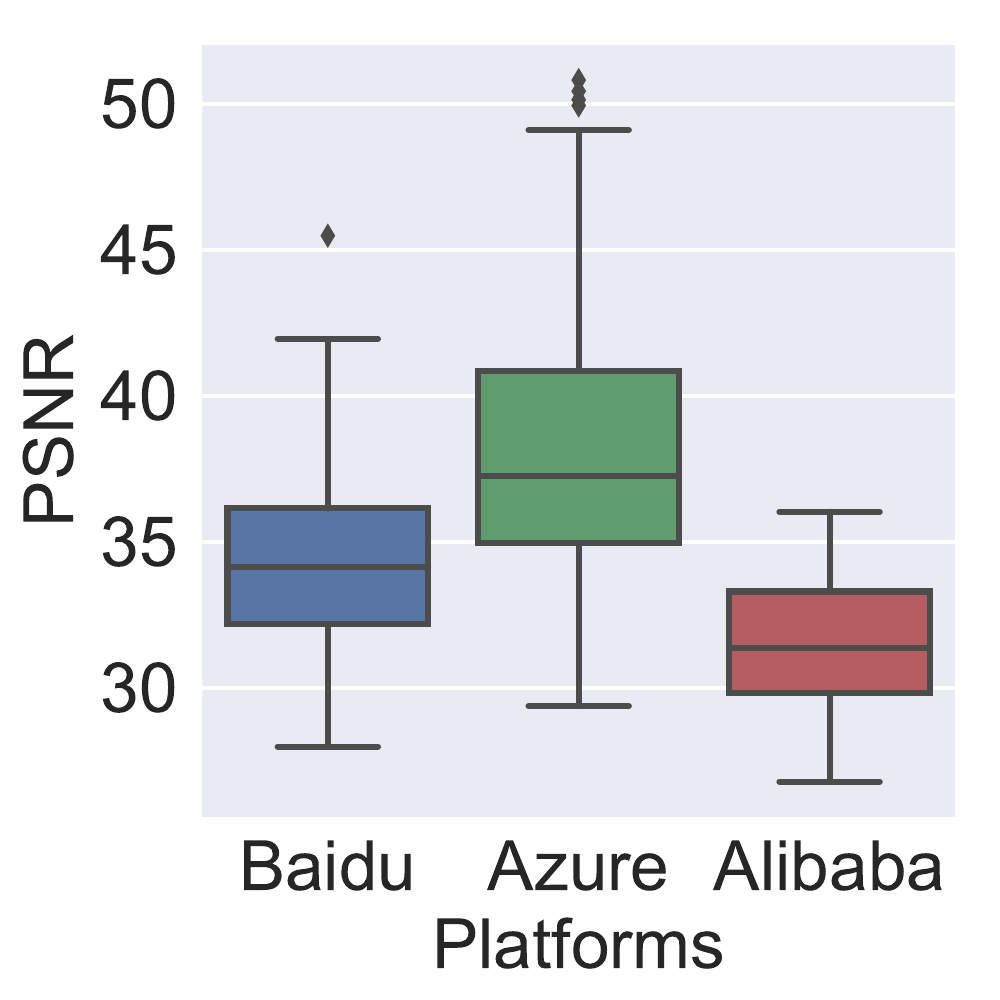}&
		\includegraphics [width=0.3\linewidth]{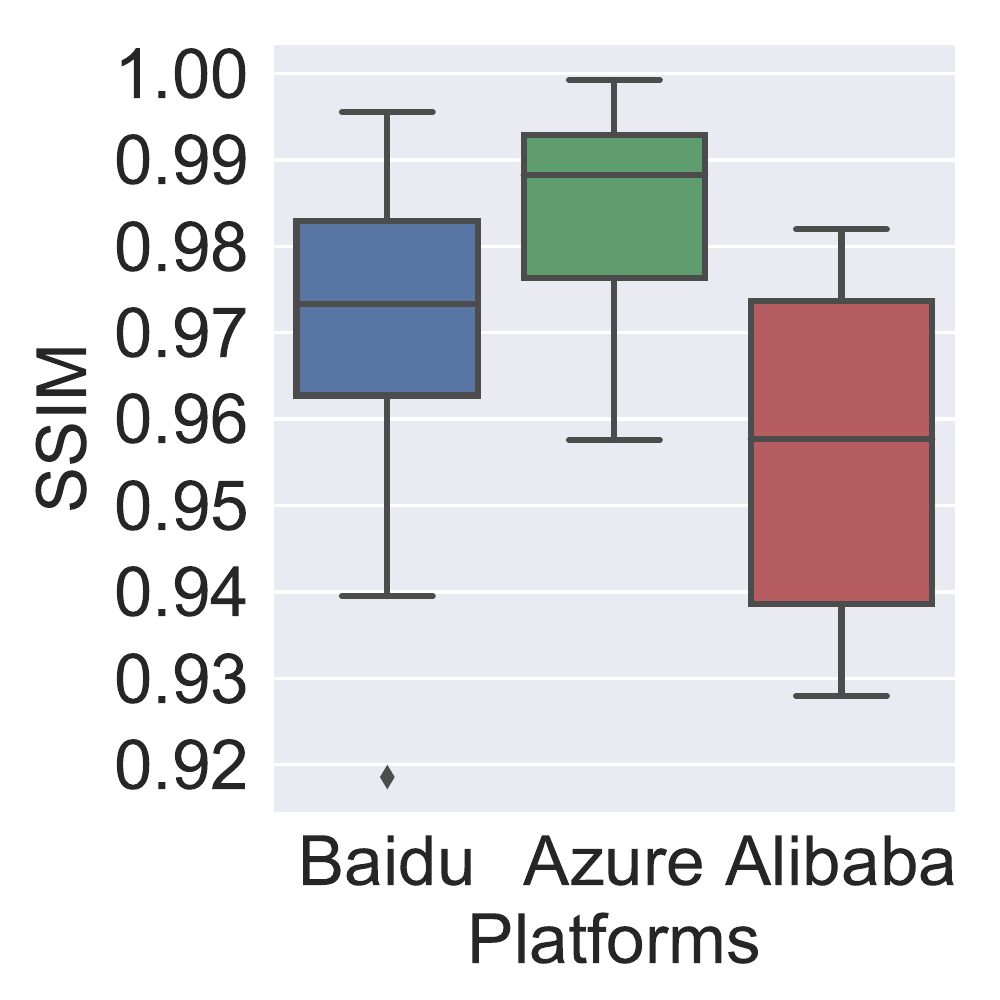}\\
		(a)&(b)&(c)\\
	\end{tabular}
	\caption{The performance of the SBLS attack on pornographic images. } \label{sex-local}
\end{figure}

The results of this attack are shown in Fig. \ref{sex-local}.
Seen from  Fig. \ref{sex-local}, Azure's detector is the weakest and \textbf{54\%} adversarial images can bypass Azure's detector.
Furthermore, since the minimum $L_0$ distance is \textbf{10}, \textbf{50} queries are sufficient for SBLS to generate an adversarial image.
Note that the prediction is \textbf{normal}, not a similar illegal category.
The success rates of attacking Baidu and Alibaba are 34\% and 19\%, respectively.
Although the success rate on Alibaba's detector is the lowest, its average $L_0$ distance is only \textbf{142}, which implies that we only modify about \textbf{0.01\%-0.6\%} pixels of the whole image ($224\times224$ pixels).
Among the successful adversarial images, all SSIM values are over 0.9 and all PSNR values are over 20.
This suggests that the quality of these adversarial images is very high.
It is easy for people to observe the pornographic information from these adversarial images.

%

\begin{figure}[!htbp]
	\centering
	\begin{tabular}{ccc}
		\includegraphics [width=0.3\linewidth]{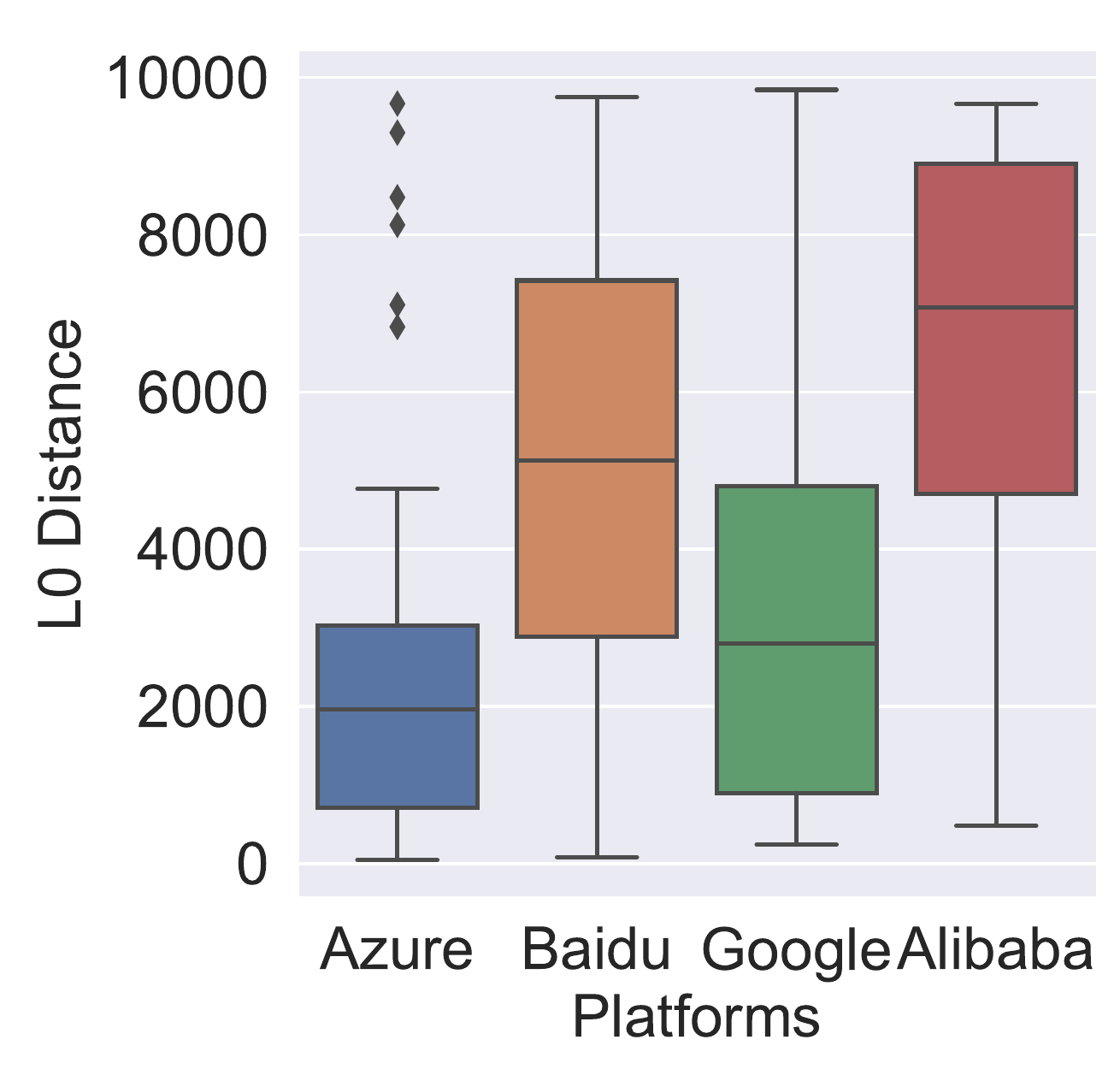} &
		\includegraphics [width=0.3\linewidth]{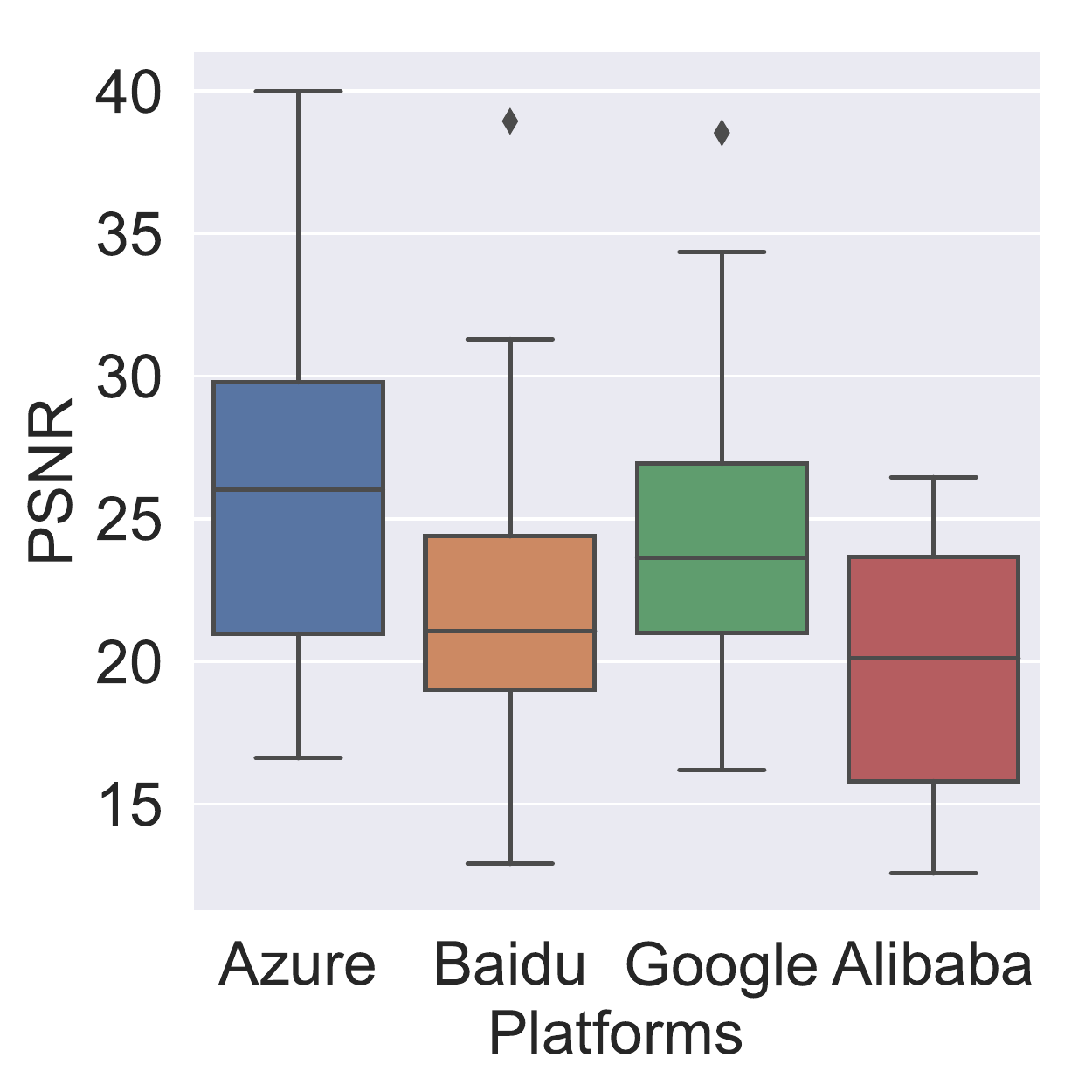}&
		\includegraphics [width=0.3\linewidth]{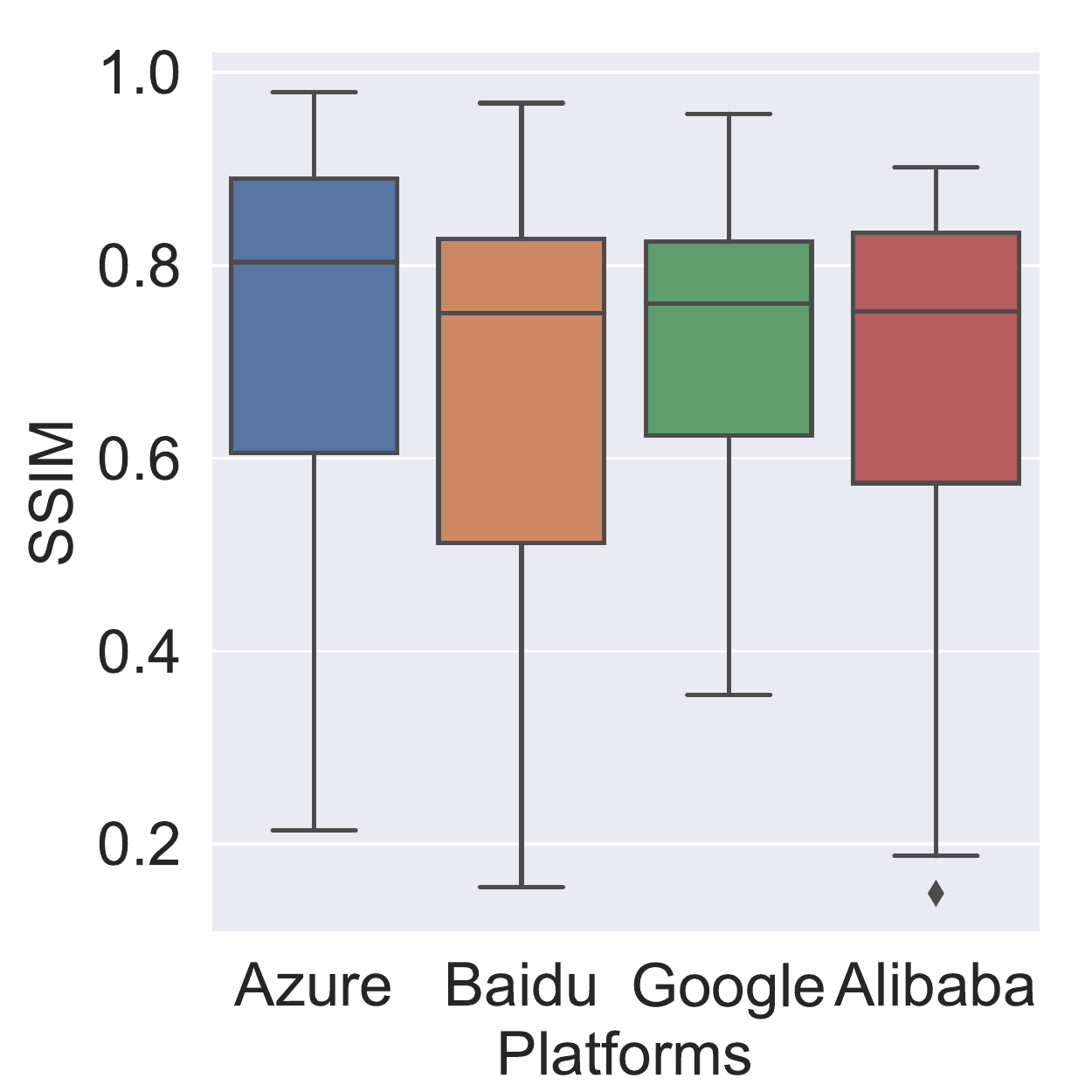}\\
		(a)&(b)&(c)\\
	\end{tabular}
	\caption{The performance of the SBB attack on pornographic images. } \label{SBBsex}
\end{figure}

 \begin{figure*}[!htbp]
	\centering
	\begin{tabular}{cc}
		\includegraphics [width=0.45\linewidth]{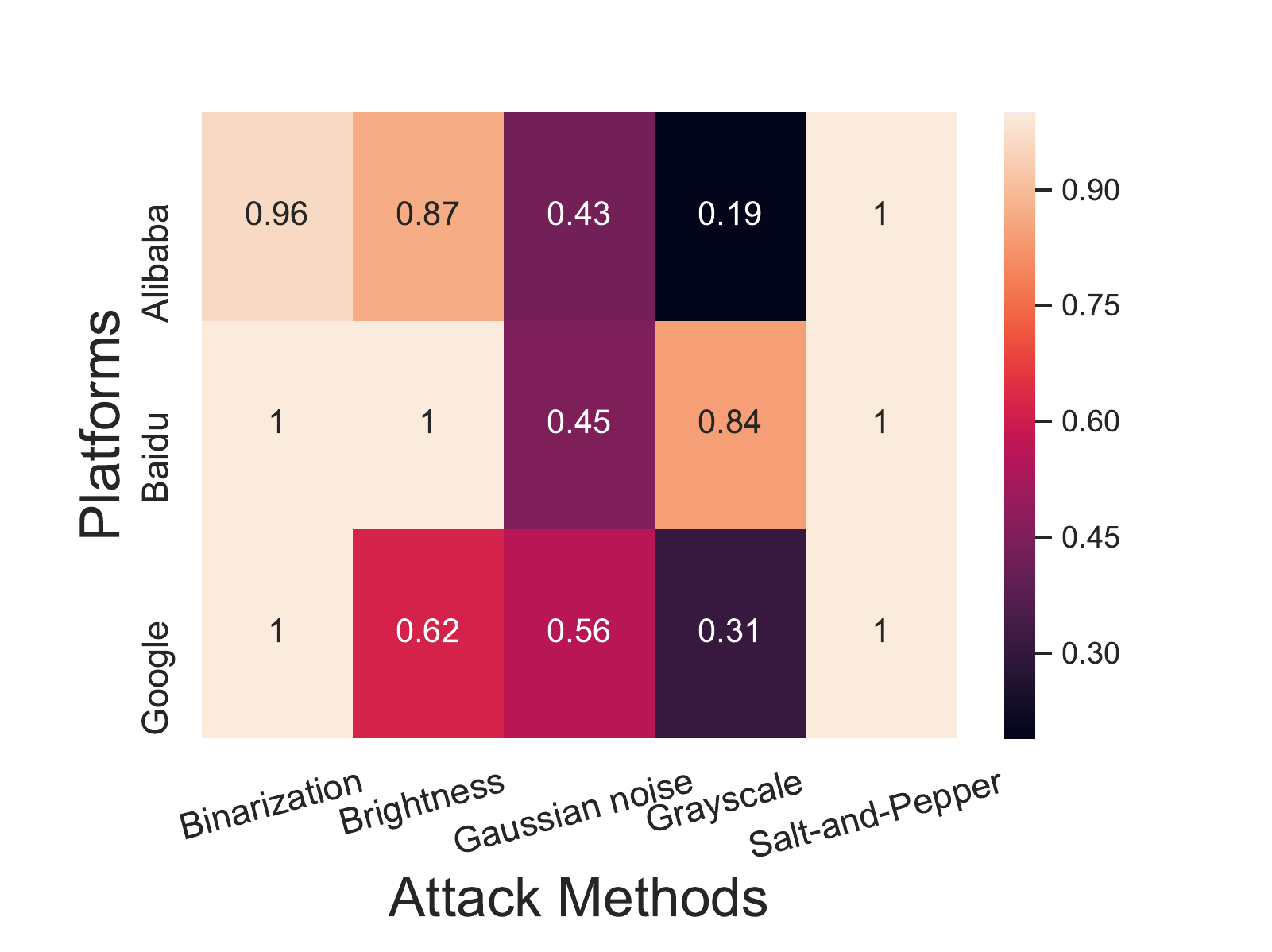} &
		\includegraphics [width=0.45\linewidth]{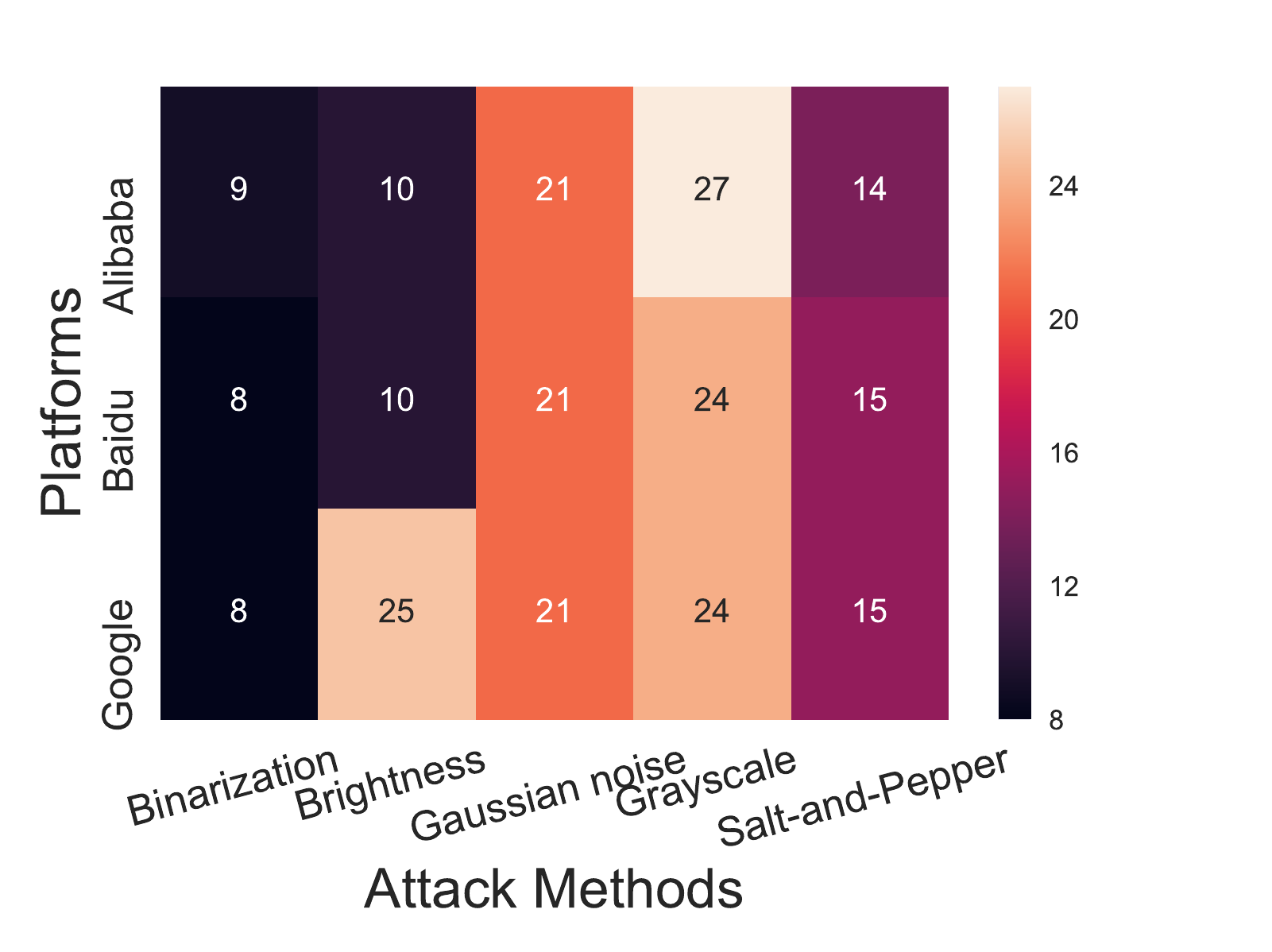}\\
		(a) Success rates &(b) PSNR values\\
	\end{tabular}
	\caption{The performance of the IP attack on violent images. } \label{violent-noise}
\end{figure*}
{\textbf{Subject-based Boundary Attack.}}
Finally, we conduct the Subject-based Boundary (SBB) attack as described in Algorithm 2.
Initially, we set $D=30$, $Step = L_0(I, adv)/10+100$.
The step size is set based on the tradeoff between accuracy and efficiency.
As the number of iterations increases, more and more pixels of the original images are recovered. 
In order to guarantee the quality of the adversarial examples, we only consider adversarial examples that recover more than 80\% of the pixels in this paper.
Finally, the success rates of attacking Azure and Google are \textbf{80\%} and \textbf{78\%}, respectively.
Besides, the success rates of attacking Baidu and Alibaba are about \textbf{40\%}.
The evaluation of these adversarial examples is shown in Fig. \ref{SBBsex}.
The successful adversarial images have lower similarity than that of the SBLS attack.
However, good adversarial examples still exist.
For Baidu's detector, the minimum $L_0$ distance is \textbf{84} and the maximum SSIM value is \textbf{0.97}.
For Azure's detector, the minimum $L_0$ distance is \textbf{48} and the maximum SSIM value is \textbf{0.98}.
The majority of PSNR values are over \textbf{20}, which means good image quality.

\subsubsection{Detectors of Violent Images}
Similarly, violent images are segmented with the semantic segmentation model FCN first.
Note that we only consider the violent images which contain persons.
In fact, a \emph{person} is a \emph{subject class} and plays an important role in the identification of violent images.
In the subsequent experiments, we focus on perturbing in the subject regions except IP attacks.

\begin{figure}[!htbp]
	\centering
	\begin{tabular}{ccc}
		\includegraphics[width=0.25\linewidth]{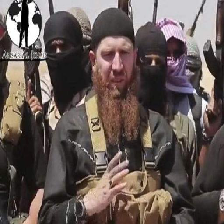}&
		\includegraphics[width=0.25\linewidth]{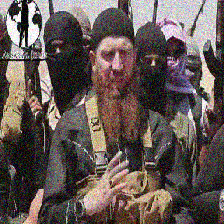}  &
		\includegraphics[width=0.25\linewidth]{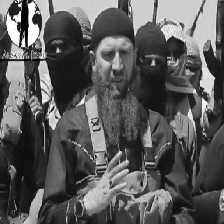}  \\
	
(a) & (b) & (c) \\
		\includegraphics[width=0.25\linewidth]{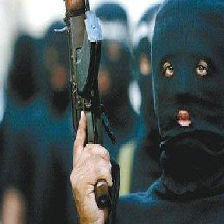}&
		\includegraphics[width=0.25\linewidth]{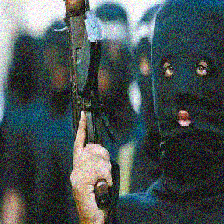}  &
		\includegraphics[width=0.25\linewidth]{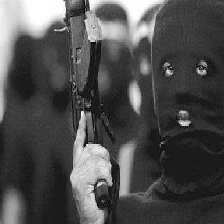}  \\
(d) & (e) & (f)\\
	\end{tabular}
	\caption{Examples on violent images: (a) and (d) are original violent images, (b) and (e) are images with Gaussian noise, and (c) and (f) are grayscale version of the original images. 
	} \label{successIPonVI}
	\vspace{-0.5em}
\end{figure}

{\textbf{Image Processing.}}
The success rates of IP attacks are shown in Fig. \ref{violent-noise} (a).
We find that the success rates of IP attacks are extremely high and the violent image detectors are easier to attack than that of pornographic image detectors.
Similar to pornographic images, Gaussian noise and Grayscale attacks can generate adversarial violent images with high quality according to their PSNR values as shown in Fig. \ref{violent-noise} (b).
We show some successful adversarial images in Fig. \ref{successIPonVI}.

{\textbf{Single-Pixel Attack.}}
\begin{figure*}[htbp]
\centering

\begin{tabular}{cc}
		\includegraphics [width=0.45\linewidth]{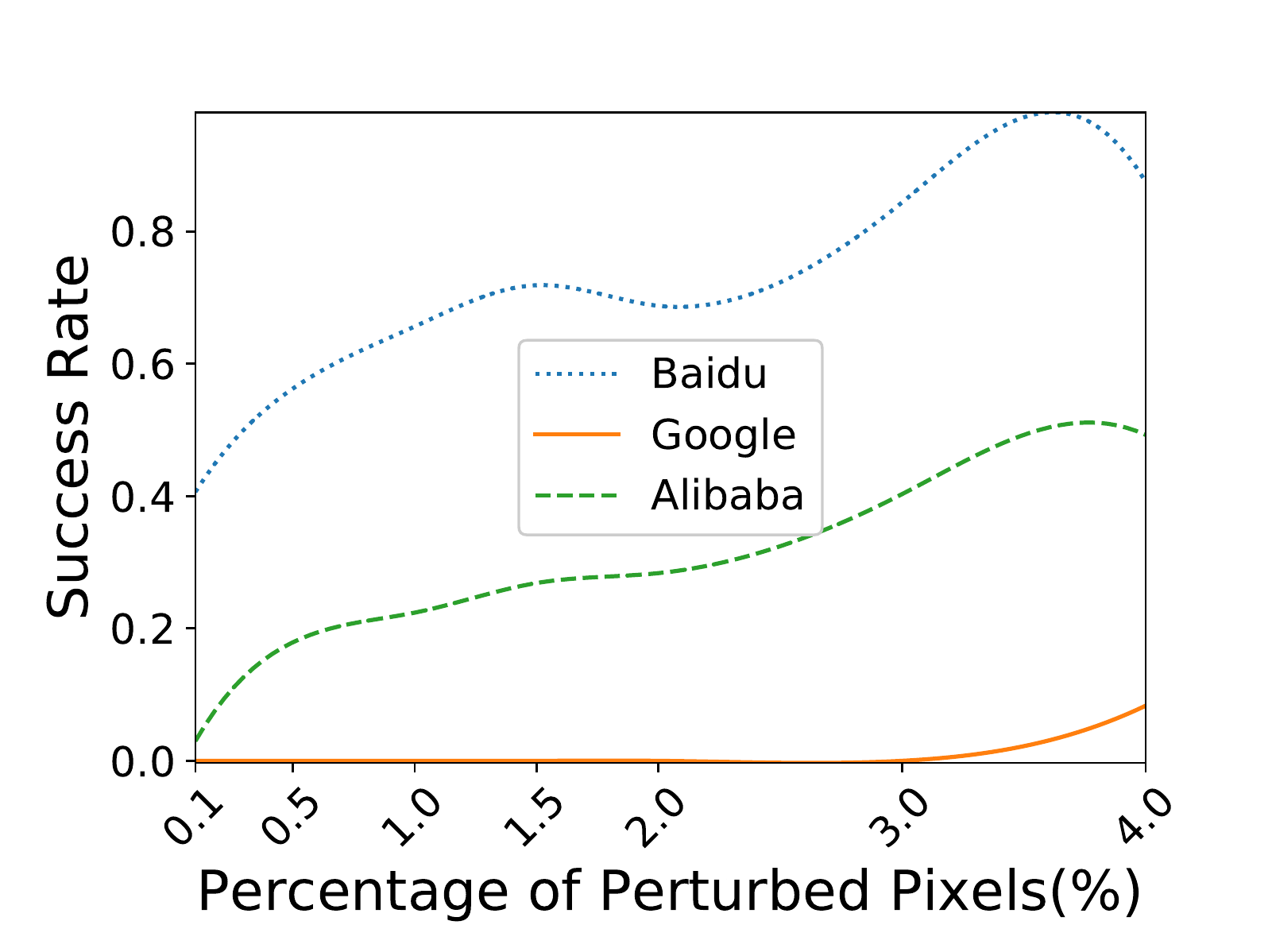}&
		\includegraphics [width=0.45\linewidth]{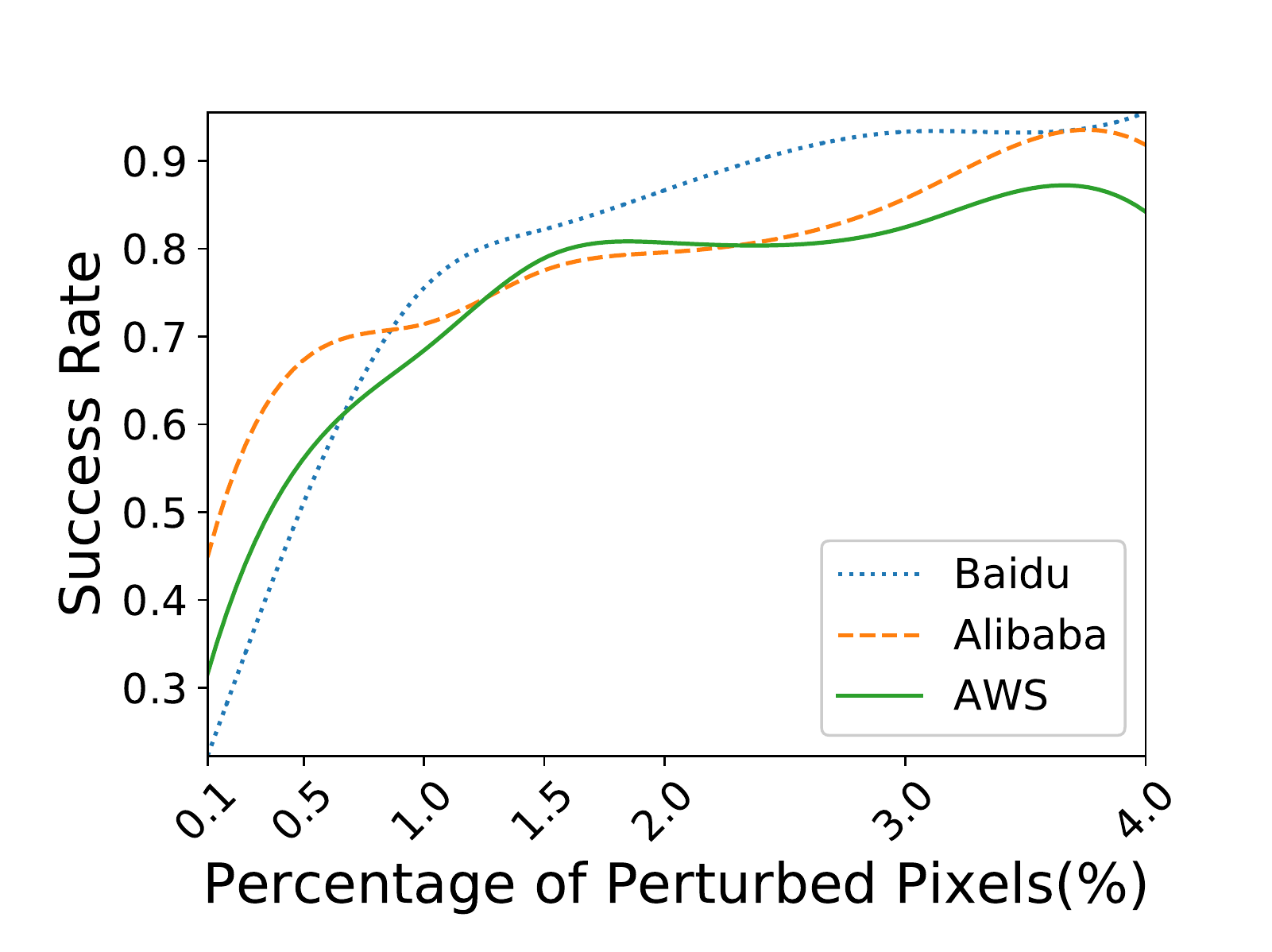}\\
		 (a) Violent images& (b) Political images\\
	\end{tabular}

\caption{SP attack on violent and political images. The horizontal axis is the number of perturbed pixels, and the vertical axis is the success rate of the attack. In the left,
the success rate of Google's detector is 0 all the time. As the number of perturbed pixels increases, more and more illegal images cannot be detected by the detectors of Alibaba and Baidu.}\label{singleviolent}
\label{fig}
\end{figure*}
The success rates of SP attacks are shown in Fig. \ref{singleviolent} (a).
To our surprise, the detectors of Baidu and Alibaba are not resistant against the SP attacks on violent images, which is very different from their performance on the pornographic images.
Google's detector shows good resilience on detecting violent images since we cannot launch a successful SP attack with small perturbation.
We speculate that different companies have different content security priorities.
For instance, Google in the United States may focus more on images filled with violence and terrorism,
while Baidu and Alibaba in China are faced with stricter censorship on pornographic images.

\begin{figure}[!htbp]
	\centering
	\begin{tabular}{ccc}
		\includegraphics [width=0.3\linewidth]{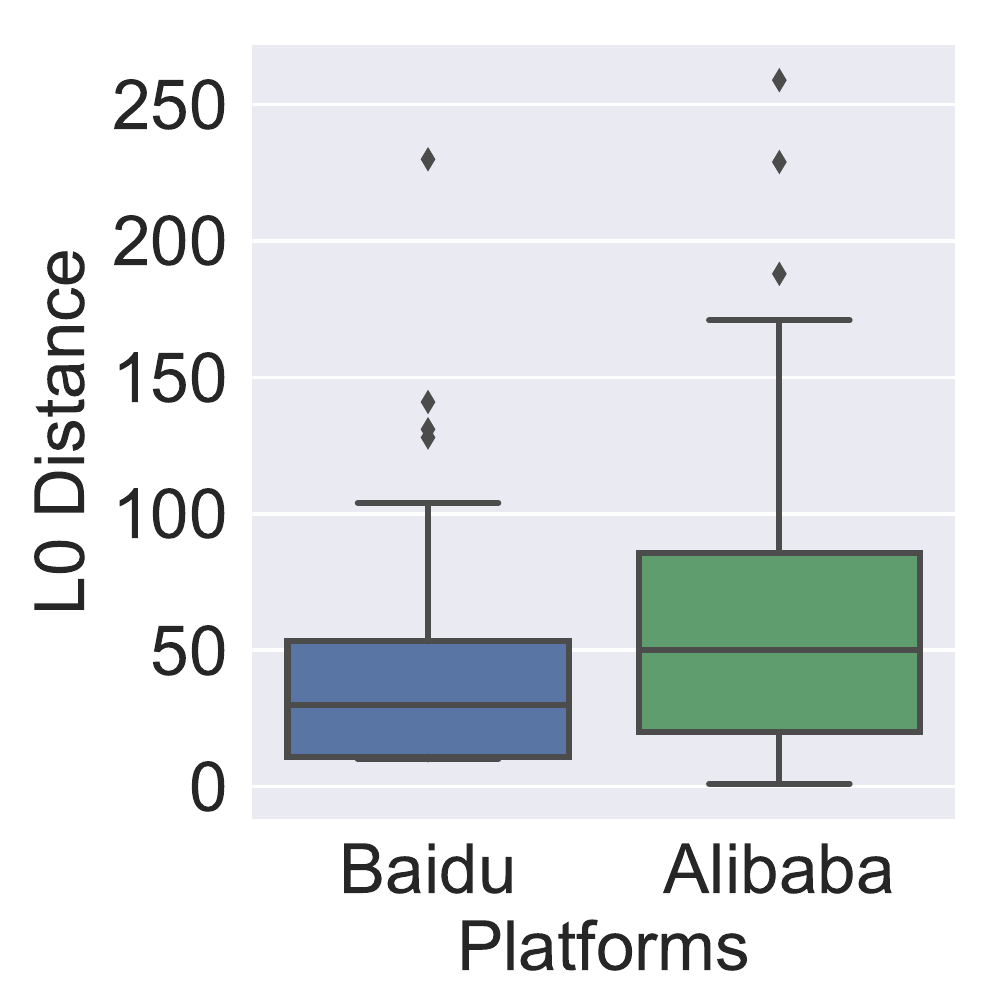} &
		\includegraphics [width=0.3\linewidth]{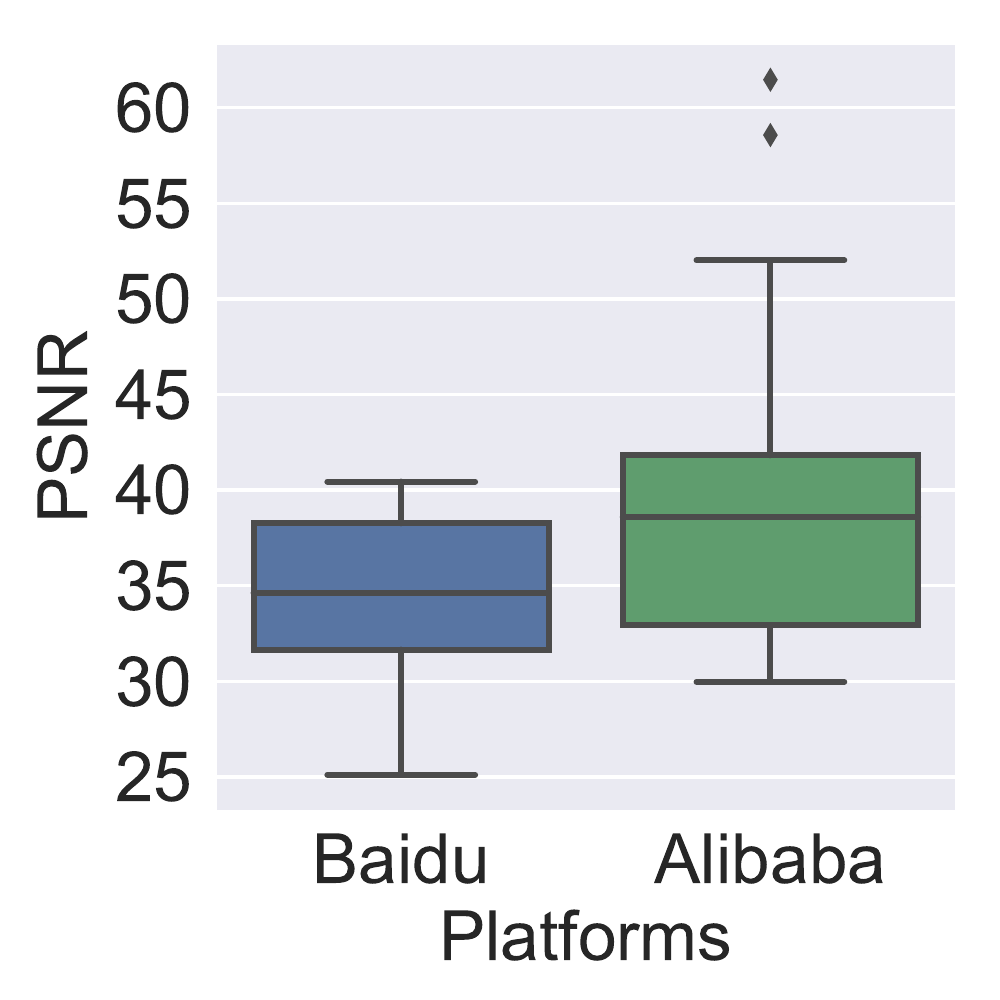}&
		\includegraphics [width=0.3\linewidth]{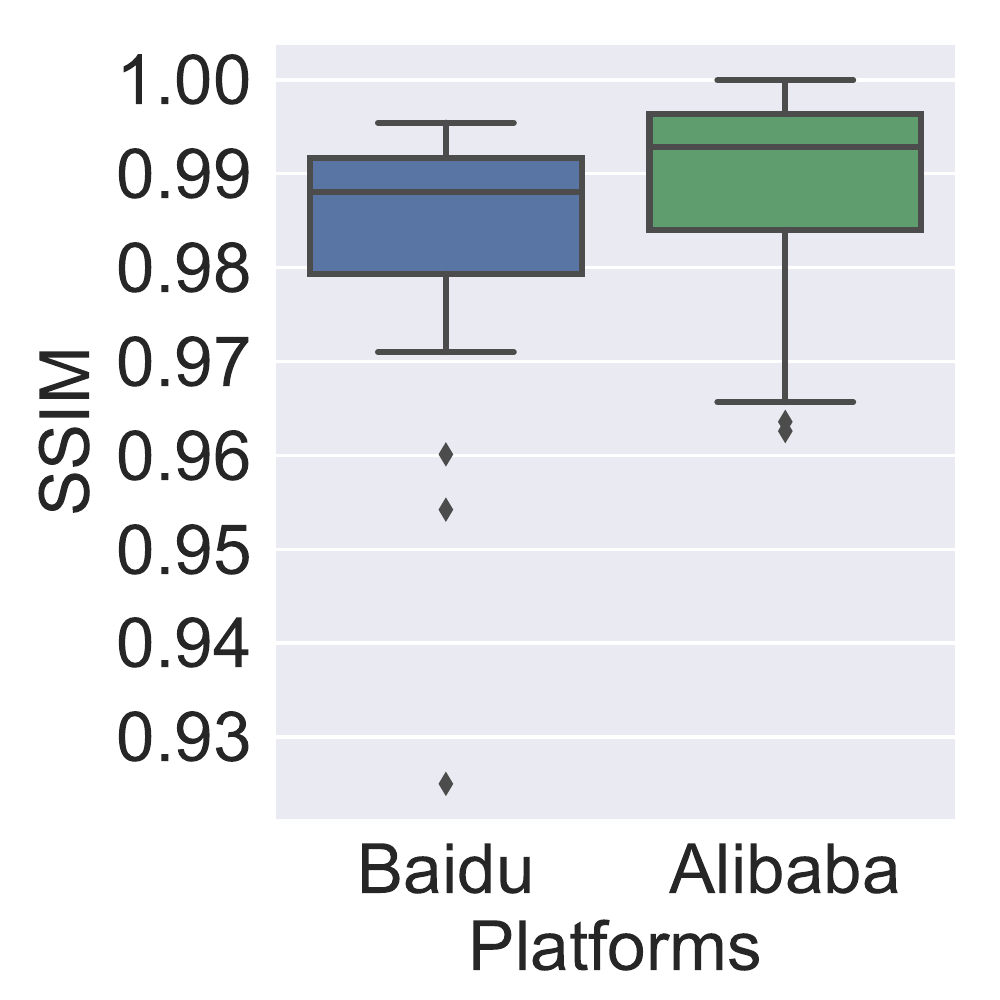}\\
		(a)&(b)&(c)\\
	\end{tabular}
	\caption{The performance of the SBLS attack on violent images.} \label{violent-local}
\end{figure}

{\textbf{Subject-based Local-search Attack.}}
Similarly, the SBLS attacks are conducted on violent images.
We set \emph{P} to 255 or 0 for different platforms and we choose the best results.
The results are shown in Fig. \ref{violent-local}.
The success rates of attacking Baidu and Alibaba are \textbf{100\%} and \textbf{72\%}, respectively, given a limited number of queries.
For instance, to successfully attack Baidu's detector, we only need 200 queries and modify 38 pixels on average. 
Besides, the average SSIM value is about \textbf{0.99} for Baidu's detector,  which reveals a high quality of the adversarial images. 

%

\begin{figure}[!htbp]
	\centering
	\begin{tabular}{cc}
		\includegraphics[width=0.4\linewidth]{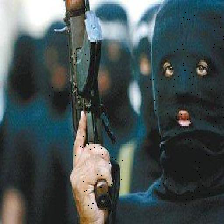}  &
		\includegraphics[width=0.4\linewidth]{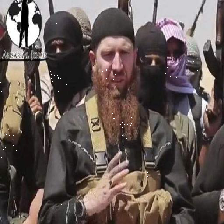}   \\
		(a)Alibaba SSIM =0.97 & (b)Baidu SSIM=0.98 \\
		PSNR=34 & PSNR =32
	\end{tabular}
	\caption{(a) is the adversarial image of Fig. \ref{successIPonVI} (d) on Alibaba's detector, and (b) is the adversarial image of Fig. \ref{successIPonVI} (a) on Baidu's detector.
	The SSIM values  and PSNR values are all pretty high, which means the perturbation is small and the similarity between the adversarial image and the original image is high.} \label{SBLSonviolent}
	\vspace{-0.5em}
\end{figure}

\begin{figure}[!htbp]
	\centering
	\begin{tabular}{ccc}
		\includegraphics [width=0.3\linewidth]{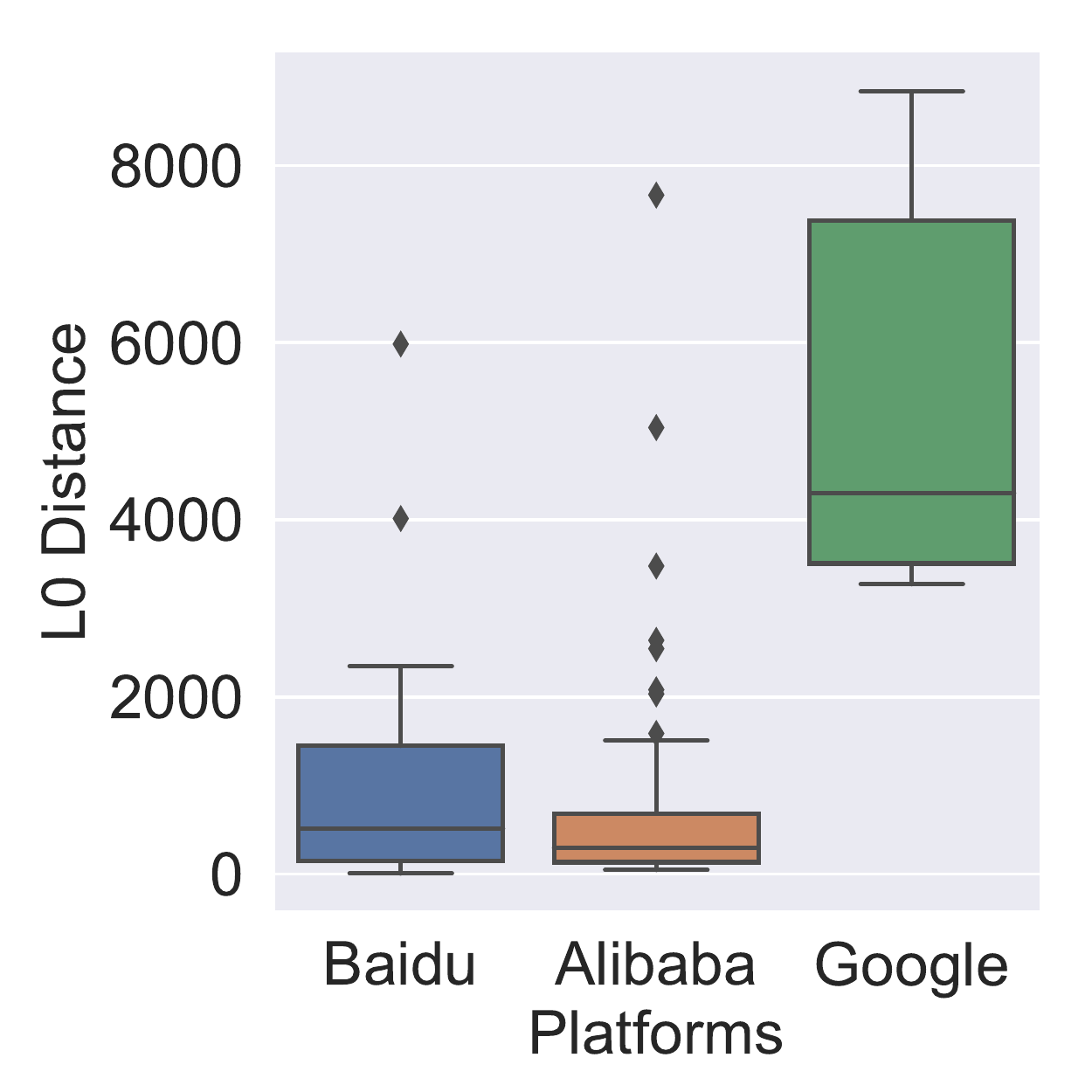} &
		\includegraphics [width=0.3\linewidth]{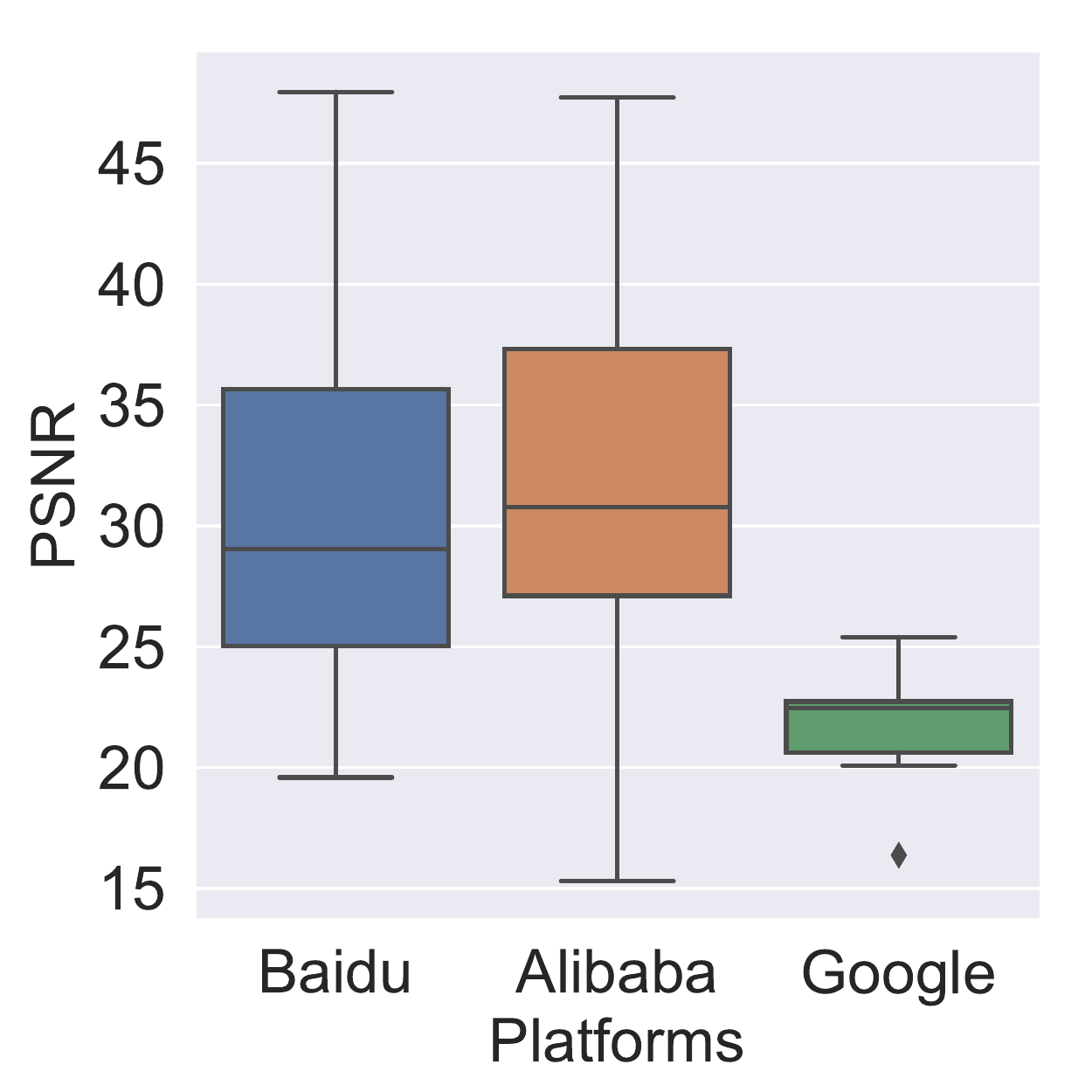}&
		\includegraphics [width=0.3\linewidth]{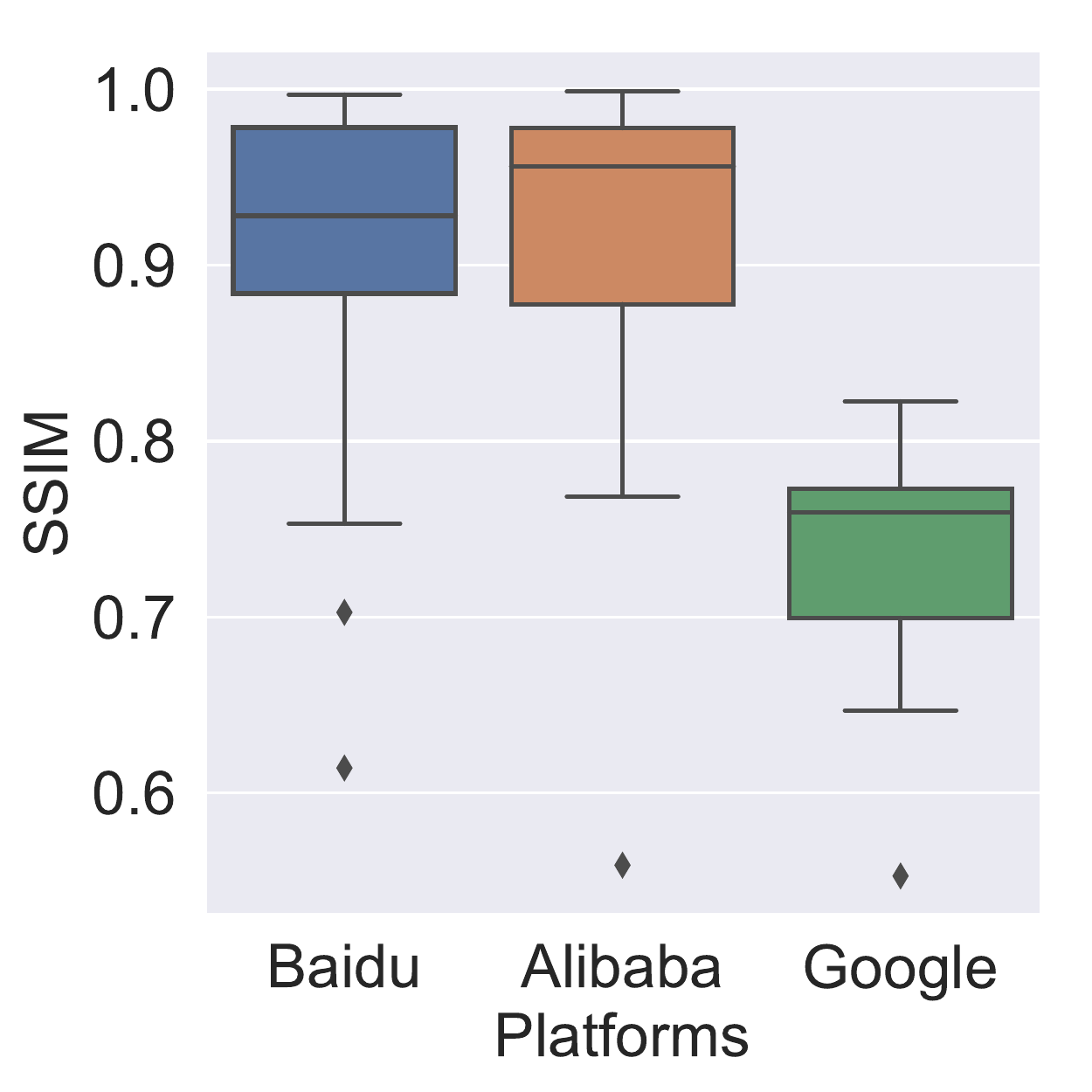}\\
		(a)&(b)&(c)\\
	\end{tabular}
	\caption{The performance of the SBB attack on violent images. } \label{SBBviolent}
\end{figure}

{\textbf{Subject-based Boundary Attack.}} We also carry out the SBB attack on violent images.
The results are shown in Fig. \ref{SBBviolent}.
The success rates of SBB attacks in all the scenarios are over \textbf{67\%}, while the quality of the adversarial images are not as good as that of SBLS, even SBB works better on violent images than pornographic images.

\begin{figure}[!htbp]
	\centering
	\begin{tabular}{ccc}
		\includegraphics[width=0.2\linewidth]{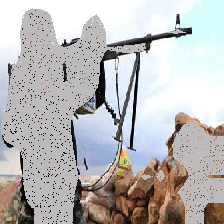}  &
		\includegraphics[width=0.2\linewidth]{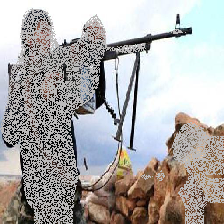} &
		\includegraphics[width=0.2\linewidth]{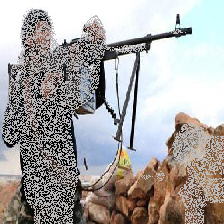}   \\
		round 1 & round 5 &round 7 \\
		
		\includegraphics[width=0.2\linewidth]{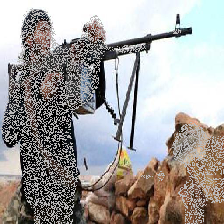}  &
		\includegraphics[width=0.2\linewidth]{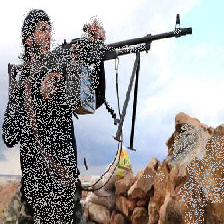} &
		\includegraphics[width=0.2\linewidth]{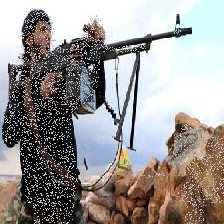}   \\
		round 9 & round 11 &round 13 \\
		
		\includegraphics[width=0.2\linewidth]{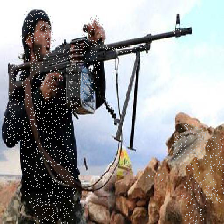}  &
		\includegraphics[width=0.2\linewidth]{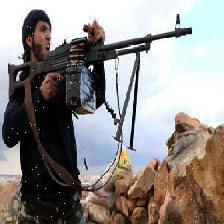} &
		\includegraphics[width=0.2\linewidth]{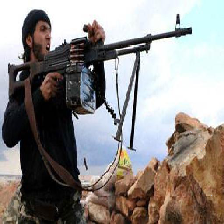}   \\
		round 15 & round 17 &round 21 \\
		
	\end{tabular}
	\caption{A successful example of SBB on Baidu's detector. In round 21, the image is still adversarial and the $L_0$ distance between it and the original image is just 10, which means
	we only perturb 10 pixels.} \label{SBBonviolent}
	\vspace{-0.5em}
\end{figure}

A successful adversarial image is shown in Fig. \ref{SBBonviolent}. 
We keep restoring the pixel until the detector recognizes it as illegal.
In round 22 the detector identifies the image as violent, and thus the image of round 21 is chosen as the final adversarial image.

\subsubsection{Detectors of Political Images}
Both Baidu and Alibaba provide the cloud service to detect whether a picture contains politicians, since sensitive political images can be insulted on the Internet.
We are interested in exploring the adversarial attacks that aim to interfere with the results from detectors of such political images.
The political images come from several countries, such as the United States, Japan, South Korea, etc.
The subject of political images is defined as a person's face since faces determine the identification of politicians.
In order to find the location of the face in the image, we adopt face detection service on open cloud platforms.
In this paper, we choose the face detection API from Baidu since it is free and easy to use.
We only need to perturb the face area rather than the entire person's pixels, which can speed up our attacks except for IP attacks.

\begin{figure}[!htbp]
	\centering
	\begin{tabular}{ccc}
		\includegraphics[width=0.3\linewidth]{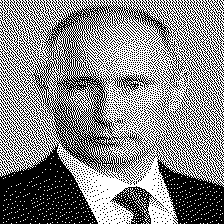}  &
		\includegraphics[width=0.3\linewidth]{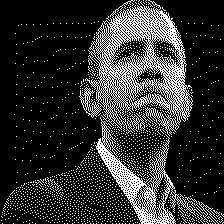}  &
		\includegraphics[width=0.3\linewidth]{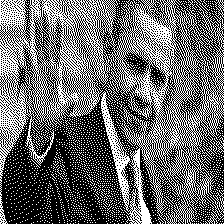} \\
		(a)Vladimir Putin & (b) Barack Obama & (c)Barack Obama\\
	\end{tabular}
	\caption{(a),(b) and (c) are all binary images. They are all labeled as normal, while their original images are all labeled as politician. }\label{successpolitics}
	\vspace{-0.5em}
\end{figure}

{\textbf{Image Processing.}}
Firstly, IP attacks are used to test political images.
We find that the success rates of Gaussian noise attacks on Baidu and Alibaba are about \textbf{76\%} and \textbf{57\%}, respectively.
For AWS's detector, its success rate is \textbf{45\%}.
\textbf{20\%} and \textbf{33\%} images can evade Baidu's and Alibaba's detection respectively using the Grayscale attack.
To our surprise, Binarization attacks perform very well.
Unlike pornographic and violent images, low-quality political images do not affect human judgment.
Several successful adversarial images of the Binarization attack are shown in Fig. \ref{successpolitics}.
Although the quality of the images is low after binarization, we can still recognize these politicians easily.

{\textbf{Single-Pixel Attack.}}
The success rates of the SP attack are shown in Fig. \ref{singleviolent} (b).
Based on the figure, political images are vulnerable to SP attacks.
Even hundreds of pixels are perturbed, the quality of images is still acceptable.

\begin{figure}[htbp]
	\centering
	\begin{tabular}{ccc}
		\includegraphics[width=0.3\linewidth]{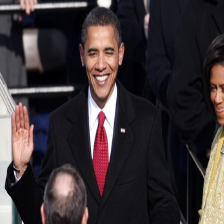}  &
		\includegraphics[width=0.3\linewidth]{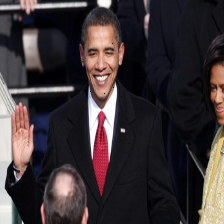}  &
		\includegraphics[width=0.3\linewidth]{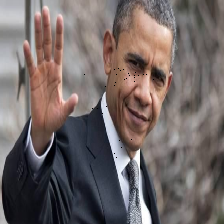}  \\
		(a)Original & (b)Adversarial & (c)Adversarial  \\
	\end{tabular}
	\caption{An example adversarial image on Baidu (b) and Alibaba (c).}\label{SBLSonpolitics}
	\vspace{-0.2em}
\end{figure}

{\textbf{Subject-based Local-Search Attack.}}
Next, we conduct the SBLS attack on political images.
The success rates of Baidu's and Alibaba's detectors are \textbf{60\%} and \textbf{46\%}, respectively, with 222 and 382 queries are made.
Here we show an example in Fig. \ref{SBLSonpolitics}.
The $L_0$ distance between Fig. \ref{SBLSonpolitics} (a) and Fig. \ref{SBLSonpolitics} (b) is \textbf{10}, which means we only perturb 10 pixels to make the image adversarial.
Among all successful examples on Baidu's detector, the average $L_0$ distance is \textbf{54}, and the average SSIM value is \textbf{0.98}.
Fig. \ref{SBLSonpolitics} (c) is an adversarial image misclassified by Alibaba's detector. Even noticeable perturbation have been added in this image, people can still
recognize Mr. Barack Obama easily through the adversarial image. In other words, if we upload the image to the cloud, the detector
will not give any warning about the politician.

{\textbf{Subject-based Boundary Attack.}}
Finally, we examine the robustness of online detectors against the SBB attack.
According to our results, we can make \textbf{82\%} images adversarial against Baidu's detector and \textbf{67\%} images adversarial against Alibaba's detector.
Moreover, the attack only needs \textbf{601} and \textbf{375} queries on average, respectively. At the meantime, a high quality can be maintained on the generated adversarial images. 

\section{ Discussion}
\subsection{Effect of Attacks}
Unlike previous attacks, our attacks do not require massive queries.
For some attack methods (e.g., IP attacks), the detectors can be bypassed without any queries.
For an iteration-based attack, only less than two thousand queries are needed to generate a good adversarial image, which can be done with the \textbf{free quota}.
Based on our investigation, almost all the major cloud service providers allow a trial period for registered users and have free invocation quotas to their APIs. 
Because of the effective attack design, all of our experiments can be completed leveraging the obtained free trials.

To achieve different goals, the proposed attacks can be launched in different scenarios. 
If the attacker cannot query the APIs frequently, the single-step IP and SP attacks can be launched.
If the attacker can freely query the APIs, they can get the adversarial examples with good quality through iterative SBLS and SBB attacks.
The success rates of IP and SP attacks are higher than that of the SBLS and SBB attacks, while the adversarial examples generated by SBLS and SBB attacks have higher quality than that of the IP and SP attacks.
The attacker can adjust the attack methods according to the actual environments.

\subsection{Defenses}
Since these attacks pose a significant threat to cloud services, it is crucial to design defense mechanisms against these attacks.
In \cite{goodfellow6572explaining} \cite{shaham2015understanding} \cite{tramer2017ensemble}, \textbf{adversarial training} is proposed to improve the robustness of deep learning models.
They iteratively create a set of adversarial examples and include them into the training data.
 Retraining a model with new training data may be helpful.
 For instance, cloud service providers can collect plenty of grayscale and binarization images or adversarial examples which are generated from the SBLS attacks and SBB attacks.
 Although adversarial training can defend against these attacks to a certain extent, obtaining adversarial examples is costly.

 In \cite{cao2017mitigating} \cite{xu2017feature}, the researchers proposed a mechanism where a detection is needed before feeding data to models.
 Certainly, \textbf{detecting inputs} can avoid some strange adversarial examples. However, this also increases increase false positives, since the input in the real world may be similar to an adversarial example.
 Besides, in \cite{carlini2017adversarial}, the authors showed that the detection mechanism can be bypassed as well.
 In \cite{xie2017mitigating} \cite{vorobeychik2014optimal}, \textbf{randomization} was proposed for mitigating adversarial effects. The models make decisions through multiple classifiers, which can increase the robustness of them. Adversarial training, detecting inputs and randomization rely on the protected model or employed training datasets, while we did not have the chance of
experimenting with them as detectors are host by third-parties.
 
Since our attacks (SBLS, SBB) rely on confidence, \textbf{rounding confidence scores} of the output is a potential good defense.
However, the outputs of APIs in our experiments are not always the exact probability values but also some scores represent confidence, which implies that, the cloud platforms may have already deployed the defense.
For instance, the outputs of AWS and Alibaba Cloud are rough scores and Google Cloud returns a word instead of a numerical representation.
Although these outputs are not precise confidence values, our attacks can still achieve high success rates, as shown in Section \ref{attackdetector}.
If the detectors output labels only, it will hinder our iterative attacks.  On the other hand, this method is not friendly to the users either and cannot help users understand the detection results.

\textbf{Limiting the number of queries} is a straightforward strategy since our SBLS and SBB attacks require iterative queries.
Specifically, we statistically calculate the successful adversarial examples of iterative attacks as shown in Table \ref{statisticsquery}.
We count the median, maximum and minimum numbers of queries required to generate adversarial examples.
In our experiments, we do not need massive queries to launch an iterative attack. The maximum number of queries is less than 3000, and over a half are no more than 1000.
Therefore, limiting the number of queries to a large threshold does not work. Besides, many websites which rely on cloud services may call the APIs multiple times within a short period to
 review the content of their webpages.

\begin{table}
\centering
\caption{The analysis of queries required for iterative attacks.}\label{statisticsquery}
\resizebox{3.5in}{!}{
\begin{tabular}{ccccccccc}
\hline

\hline

\hline
&& \multicolumn{3}{c}{SBLS}  &\multicolumn{3}{c}{SBB}\\
Type&Platform&max&median&min&max&median&min\\
\hline 

\hline 

\hline 
\multirow{4}{*}{Porn}& Google&-&-&-&1080&660&180\\
& Azure&550&50&50&1500&990&240\\
&Baidu&1350&350&100&2820&1710&210\\
&Alibaba&1302&618&152&330&240&150\\
\hline

\hline 
\multirow{3}{*}{Violence}& Google&-&-&-&660&540&240\\
&Baidu&1100&150&50&1500&780&90\\
&Alibaba&1301&251&3&1080&780&210\\
\hline

\hline 
\multirow{2}{*}{Politician}&Baidu&650&250&50&1650&690&60\\
&Alibaba&784&101&4&900&360&60\\

\hline

\hline

\hline

\end{tabular}
}
\end{table}

\begin{table*}
\centering
\caption{The comparison of IP attacks under different noise filters.}\label{noisefilters}
{
\begin{tabular}{cccc|ccc|c|ccc|cc}
\hline

\hline

\hline
& &\multicolumn{2}{c|}{AWS}&\multicolumn{3}{c|}{Alibaba}&Azure&\multicolumn{3}{c|}{Baidu}&\multicolumn{2}{c}{Google}\\
& &porn&politician&porn&politician&violence&porn&porn&politician&violence&porn&violence\\
\hline
\multirow{3}{*}{Binarization}&Original&0.98&0.99&0.88&0.9&0.96&0.88&1&1&1&1&1\\
&Gaussian Filter&0.79&0.68&0.48&0.59&0.73&0.16&0.22&0.68&1&1&0.68\\
&Median Filter&0.98&0.94&0.88&0.9&0.88&0.88&1&0.98&1&1&0.87\\
\cline{2-13}
\multirow{3}{*}{Brightness}&Original&1&0.88&0.96&1&0.87&1&0.55&1&1&0.99&0.62\\
&Gaussian Filter&0.98&0.72&0.98&0.95&0.52&0.34&0.63&0.87&0.91&0.97&0.62\\
&Median Filter&0.94&0.81&0.97&1&0.55&0.3&0.7&0.92&1&0.97&0.68\\
\cline{2-13}
\multirow{3}{*}{Gaussian noise}&Original&0.46&0.45&0.03&0.57&0.43&0.56&0.02&0.76&0.45&0.54&0.56\\
&Gaussian Filter&0.22&0.35&0.1&0.51&0.09&0.2&0.05&0.55&0.2&0.38&0.13\\
&Median Filter&0.37&0.44&0.42&0.53&0.12&0.4&0.04&0.62&0.28&0.43&0.18\\
\cline{2-13}
\multirow{3}{*}{Grayscale}&Original&0.11&0.07&0&0.33&0.19&0.22&0&0.2&0.84&0.04&0.31\\
&Gaussian Filter&0.11&0.05&0.22&0.31&0.13&0.1&0.04&0.22&0.35&0.04&0.43\\
&Median Filter&0.07&0.06&0.22&0.31&0.13&0.16&0.03&0.33&0.55&0.04&0.31\\
\cline{2-13}
\multirow{3}{*}{Salt-and-Pepper}&Original&1&1&1&1&1&1&1&1&1&1&1\\
&Gaussian Filter&0.98&0.83&0.61&0.98&0.99&1&1&1&1&1&0.63\\
&Median Filter&0.08&0.26&0.07&0.46&0.1&1&0.9&1&1&1&0.61\\

\hline

\hline

\hline

\end{tabular}
}
\end{table*}

In \cite{hosseini2017google}, the authors claimed that \textbf{noise filters} could be an effective method to eliminate the adversarial attacks.
In order to understand whether the noise filter is effective, we implement it offline and the noise filter only processes the images sent to the cloud-based detectors.
In our experiments, we select the Gaussian Filter and Median Filter since the Gaussian filter can handle Gaussian noise and the Median Filter can handle Salt-and-Pepper noise.
For Binarization, Brightness, and Grayscale attacks, we also use both the Gaussian filter and Median Filter to process them since there is not a specific filter to handle them.
To test IP attacks, we record the success rates of attacks with or without noise filters.
The results are shown in Table \ref{noisefilters}.
From Table \ref{noisefilters}, we can clearly see a decline in the success rates of Gaussian noise attacks and Salt-and-Pepper attacks. 
However, for Azure, Baidu, and Google, the success rates of Salt-and-Pepper attacks remain high even under the Gaussian Filter.
We analyze these adversarial examples and find that the perturbation parameters of Salt-and-Pepper noise are almost twice as large as the original adversarial examples'.
Besides, the Gaussian Filter is more effective than the Median Filter to defend against Binarization attacks.
However, the Gaussian Filter and Median Filter affect the accuracy of the detection models to a certain extent.

As for SP, SBLS and SBB attacks, they can be considered as adding noise to the images. Therefore, the Gaussian Filter and Median Filter are also selected in our experiments.
The experimental results show that the Gaussian Filter has almost no impact on the SP attacks. The Median Filter can reduce the success rates of SP attacks on pornography to less than 5\%, and cut the success rates of  SP attacks on violent and political images in half. Through the observation of filtered images, we think the Median Filter allows pornographic images to show more pixels of the skin color, making them easier for detection, while the elimination of random noise from violent and political images also increases their blurring.
To simulate noise filters on cloud platforms, noise filters are adopted in each iteration of SBLS and SBB.
Then the final success rates of the attacks are recorded.
During our experiments, we find that the Gaussian Filter has little impact on either SBLS or SBB attacks.
When using the Median Filter, SBLS attacks cannot work since they try to capture the difference of the prediction caused by small perturbations in each iteration, while the Median filter can easily filter out small perturbations. For SBB attacks, the Median Filter could be able to reduce the success rates by approximately 10\%.
The reason is that the initial perturbation of SBB attacks is large, and the Median Filter does not work well for large perturbations in the region.
Nevertheless, with the deployment of noise filters, we need more perturbations to launch successful attacks.

In summary, we find that the Median Filter can resist most Salt-and-Pepper noise attacks, SP attacks, and SBLS attacks. The Gaussian filter can resist most Gaussian noise attacks. 
However, it is difficult for cloud-based detectors to deploy a uniform filter and thus it is difficult to simultaneously defend against all kinds of attacks. In addition, the noise filters will also introduce a decrease in model accuracy.

\section{Related Work}

Previous works mainly study the security and privacy of deep neural networks under white-box models \cite{szegedy2013intriguing} \cite{papernot2016limitations} \cite{goodfellow6572explaining} \cite{moosavi2016deepfool} \cite{carlini2016towards} \cite{papernot2015distillation}.
In the white-box setting, the attacker can obtain the adversarial examples quickly and accurately. Besides, the perturbation is small.
However, it is usually difficult for an attacker to know the inner parameters of a model in the real world. For instance, the architecture and parameters of deep models on cloud platforms
cannot be obtained. The attacker can only access the APIs opened by cloud platforms. Thus, the black-box attack of neural networks is more threatening.

Researchers have launched some black-box attacks on deep neural networks recently.
 In \cite{papernot2017practical}, Papernot et al. proposed that the attacker can train a substituted model, which approaches the target model, and then generate
adversarial examples on the substituted model. Their experiments showed that good transferability exists in adversarial examples, while the attack is not totally black-box.
They have knowledge of the training data and test the attack with the same distributed data.
In \cite{liu2016delving}, Liu et al. adopted an ensemble-based model to improve transferability and successfully attacked Clarifai.com.
However, the classifiers are greatly different from detectors.  
For complicated detectors, transferability is not very well.

The query-based black-box attack has also been explored by researchers.
In \cite{tramer2016stealing} \cite{shokri2017membership} \cite{fredrikson2015model}, researchers can get inner information of models through lots of queries,
which may be impractical in real applications.
In \cite{ilyas2017query}, thousands of queries are made for low-resolution images to generate adversarial examples.
For high-resolution images, it still takes tens of thousands of queries, which is also impractical.
By querying the output of the target model, gradient estimation based black-box attack methods were proposed in \cite{chen2017zoo} \cite{bhagoji2017exploring}.
Nevertheless, faced with high-resolution images,
millions of queries are required, which is very inefficient and impractical.
In \cite{brendel2017decision}, Brendel et al. proposed a decision-based attack that starts from a large adversarial
perturbation and then seeks to reduce the perturbation while staying adversarial. However, it takes 1,200,000 queries on average to generate a good adversarial example
for high-resolution images. In this paper, we utilize the semantic segmentation technique to speed up the process. Besides, we choose the important pixels according to
the returned probability.
In \cite{shiva2017simple}, Nina et al. proposed a greedy local search algorithm to attack black-box models.
The SBLS attack in this paper is originated from \cite{shiva2017simple}, and we have extended it to cloud platforms.

Several other forms of black-box attacks are studied in recent works. In \cite{athalye2017synthesizing}, Anish et al. constructed real-world 3D objects that consistently
fool neural networks across a wide distribution of angles and viewpoints. They showed that adversarial
examples are a practical concern for real-world systems. In \cite{hosseini2017google}, Hossein et al. found that an API generates completely different outputs for the noisy images, while a human
observer would perceive its original content.
We expand the experiment and use five image processing techniques to attack models of five cloud platforms.
For detectors, we find Salt-and-Pepper noise is not valid enough and other image processing methods may perform well.
In \cite{oh2017whitening}, Seong et al. inferred the inner information of models through multiple queries and the revealed internal
information helps generate more effective adversarial examples against a black-box model. Aiming at this, thousands of models should be trained.
However, this is impractical and it is difficult to have multiple candidate models to infer when handling detectors.

\section{Conclusion and Future Work}
In this study, we conduct a comprehensive security study for the cloud-based image detectors.
We design four kinds of attacks and verify them on major cloud service platforms.
According to our experimental results, we find that cloud-based detectors are easily bypassed.
We reported our findings to the tested cloud platforms and received positive feedbacks from them.

In the future, 
we are planning to further improve the effectiveness of our algorithms.
In addition, a general method to attack all platforms with a small number of queries would be a meaningful topic.
It is also important to design the defense mechanisms for these cloud services against adversarial example attacks.
For instance, cloud platforms can perform effective detection before outputting a label and distinguish between malicious tests and normal samples.
We hope our work can help cloud platforms design secure services and provide inspiration to researchers in the deep learning security area.

\bibliographystyle{unsrt}
\bibliography{refs}

\end{document}